%% file: t-viSNE.tex
\documentclass[10pt,journal,compsoc]{IEEEtran}
\usepackage[absolute,overlay]{textpos}
\usepackage{pbox}
\usepackage{everypage}

\newcommand{\stamp}[1][© 2020 IEEE. This is the author's version of the article that has been published in IEEE Transactions on Visualization and Computer Graphics. The final version of this record is available at: \href{https://doi.org/10.1109/TVCG.2020.2986996}{\color{blue}10.1109/TVCG.2020.2986996}]{%
\begin{textblock*}{165mm}(25mm,270mm)
\centering%
\small%
\emph{#1}%
\end{textblock*}%
}

\AddEverypageHook{
  \stamp
}

\usepackage{caption}
\usepackage{amsmath}

\ifCLASSOPTIONcompsoc
  \usepackage[nocompress]{cite}
\else
  \usepackage{cite}
\fi
\ifCLASSINFOpdf
   \usepackage[pdftex]{graphicx}
   \graphicspath{{figures/}{pictures/}{photos/}{images/}{./}}
   \DeclareGraphicsExtensions{.pdf,.png,.jpg,.jpeg}
\else
   \usepackage[dvips]{graphicx}
   \DeclareGraphicsExtensions{.eps}
\fi

\usepackage{mathptmx}                  
\usepackage[hidelinks]{hyperref}

\usepackage[dvipsnames]{xcolor} 

\newcommand{\hl}[1]{#1} 
\newcommand{\deleted}[1]{} 

\newcommand{\cit}[1]{``#1''}
\newcommand{\circled}[1]{\raisebox{.5pt}{\textcircled{\raisebox{-.9pt} {#1}}}}

\usepackage{csquotes}
\usepackage{cleveref}
\renewcommand{\autoref}{\Cref}

\newcommand{\mypar}[1]{\vspace{.5em} \noindent \textbf{#1} \quad}

\begin{document}
%
\title{t-viSNE: Interactive Assessment and Interpretation of t-SNE Projections}
%
%
%
%

\author{Angelos~Chatzimparmpas,~\IEEEmembership{Student~Member,~IEEE,}
        Rafael M.~Martins,~\IEEEmembership{Member,~IEEE~Computer~Society,}
        and~Andreas~Kerren,~\IEEEmembership{Senior~Member,~IEEE}
\IEEEcompsocitemizethanks{\IEEEcompsocthanksitem Angelos Chatzimparmpas, Rafael M. Martins, and Andreas Kerren are with the Department of Computer Science and Media Technology, Linnaeus University, V{\"a}xj{\"o} 35195, Sweden. \protect\\
E-mail: \{angelos.chatzimparmpas,rafael.martins,andreas.kerren\}@lnu.se.}
\thanks{Manuscript received October XX, 2019; revised February XX, 2020.}}

%
%

\markboth{IEEE TRANSACTIONS ON VISUALIZATION AND COMPUTER GRAPHICS,~Vol.~XX, No.~X, JULY~2020}%
{Chatzimparmpas \MakeLowercase{\textit{et al.}}: t-viSNE: \\ Interactive Assessment and Interpretation of t-SNE Projections}
\IEEEtitleabstractindextext{%
\begin{abstract}
\input{0.Abstract}
\end{abstract}

\begin{IEEEkeywords}
Interpretable t-SNE, dimensionality reduction, high-dimensional data, explainable machine learning, visualization.
\end{IEEEkeywords}}

\maketitle

  \stamp

\IEEEdisplaynontitleabstractindextext

%
\IEEEpeerreviewmaketitle

\IEEEraisesectionheading{\section{Introduction} \label{intro}}%
  \input{1.Introduction}

\section{Related Work} \label{sec:relwork}%
  \input{2.Related_Work}
\section{Overview of t-SNE} \label{sec:tsne}
  \input{3.t-SNE}


\section{t-viSNE: A Visual Inspector of t-SNE} \label{t-viSNE}%
  \input{4.System_Overview}
\section{Use Cases} \label{case}%
  \input{5.Case_Study}
\section{User Evaluation} \label{eval}%
  \input{6.Evaluation}

\section{Discussion} \label{sec:disc}%
  \input{7.Discussion}

\section{Conclusions} \label{conclus}%
  \input{8.Conclusion}
\ifCLASSOPTIONcompsoc
  \section*{Acknowledgments}
  The authors are thankful to Margit Pohl, Vienna University of Technology, for her suggestions to improve the evaluation section.
\else
  \section*{Acknowledgment}
\fi





%



\bibliographystyle{IEEEtran}
\bibliography{references}

%
\vskip -2.5\baselineskip plus -1fil
\begin{IEEEbiography}[{\includegraphics[width=1in,height=1.25in,clip,keepaspectratio]{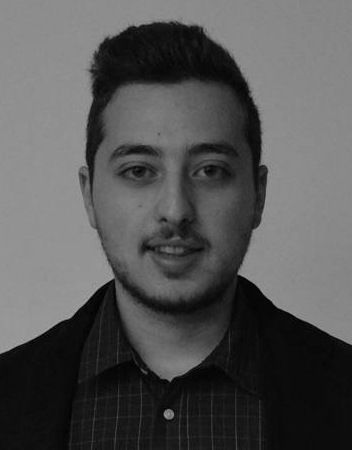}}]{Angelos Chatzimparmpas}
is PhD student within the ISOVIS research group and the Linnaeus University Centre for Data Intensive Sciences and Applications at the Department of Computer Science and Media Technology, Linnaeus University, Sweden. His main research interests include visual exploration of the inner parts and the quality of machine learning models with a specific focus on engineering smarter cyber-physical systems, as well as visual analytics approaches involving such models.
\end{IEEEbiography}
\vskip -2\baselineskip plus -1fil
\begin{IEEEbiography}[{\includegraphics[width=1in,height=1.25in,clip,keepaspectratio]{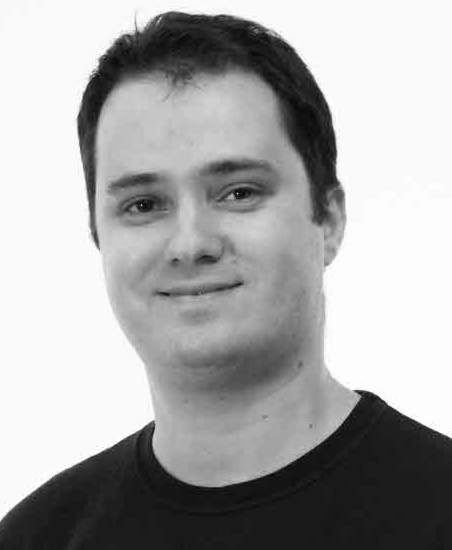}}]{Rafael M. Martins}
is Senior Lecturer at the Department of Computer Science and Media Technology at Linnaeus University, Sweden. His PhD research involved mainly the visual exploration of the quality of dimensionality reduction (DR) techniques, a topic he continues to investigate, in addition to other related research areas such as the interpretation of DR layouts and the application of DR techniques in different domains including software engineering and digital humanities.
\end{IEEEbiography}
\vskip -2\baselineskip plus -1fil
\begin{IEEEbiography}[{\includegraphics[width=1in,height=1.25in,clip,keepaspectratio]{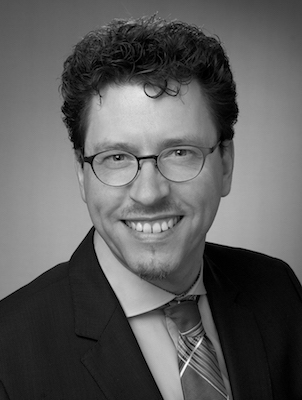}}]{Andreas Kerren}
is Professor of Computer Science at the Department of Computer Science and Media Technology at Linnaeus University, Sweden, and head of the ISOVIS research group. He is also a key researcher at the Linnaeus University Centre for Data Intensive Sciences and Applications contributing with his expertise in information visualization and visual analytics. His research mainly focuses on the explorative analysis and visualization of typically large and complex information spaces, for example in the humanities or the life sciences.
\end{IEEEbiography}
\vskip -2\baselineskip plus -1fil
\vfill



\end{document}

%% file: 0.Abstract.tex
t-Distributed Stochastic Neighbor Embedding (\mbox{t-SNE}) for the visualization of multidimensional data has proven to be a popular approach, with successful applications in a wide range of domains. Despite their usefulness, t-SNE projections can be hard to interpret or even misleading, 
which hurts the trustworthiness of the results. Understanding the details of t-SNE itself and the reasons behind specific patterns in its output may be a daunting task, especially for non-experts in dimensionality reduction. 
In this work, we present t-viSNE, an interactive tool for the visual exploration of t-SNE projections that enables analysts to inspect different aspects of their accuracy and meaning, such as the effects of hyper-parameters, distance and neighborhood preservation, densities and costs of specific neighborhoods, and the correlations between dimensions and visual patterns. We propose a coherent, accessible, and well-integrated collection of different views for the visualization of t-SNE projections. The applicability and usability of t-viSNE are demonstrated through hypothetical usage scenarios with real data sets. Finally, we present the results of a user study where the tool's effectiveness was evaluated. By 
bringing to light information that would normally be lost after running t-SNE, we hope to support analysts in using t-SNE 
and making its results better understandable.

%% file: 1.Introduction.tex
\IEEEPARstart{D}{imensionality} Reduction (DR) techniques are an important part of the toolbox of high-dimensional data analysis, with its initial techniques such as Principal Component Analysis (PCA)~\cite{jolliffe16} and Multidimensional Scaling (MDS)~\cite{borg2005modern} being several decades old now.
The problem that DR tries to solve is, in general, to find a low-dimensional representation of a high-dimensional data set that retains---as much as possible---its original \emph{structure}. 
%
When used for visualization, the output is set to two or three dimensions, and the results are commonly visualized with scatterplots, where similar objects are modeled by nearby points, and dissimilar ones by distant points.

\emph{Linear DR methods}, such as PCA, are easier to understand and to explain, since the remaining axes are linear combinations of the original dimensions, which establishes a direct relationship between the low-dimensional and the high-dimensional data set. 
%
When the specific constraints of being simple and easily explainable are abrogated, other more intricate \emph{non-linear DR} (or \emph{manifold learning}) \emph{methods} can be used in order to capture much more complex high-dimensional patterns~\cite{lee07}. In general, non-linear DR methods opt to maintain \emph{local} structures in detriment of \emph{global} ones, i.e., their algorithms favor the optimization of neighborhoods of points and mostly disregard large distances.
Although non-linear DR methods have also been around for quite some time (e.g., Sammon Mapping~\cite{sammon69}), they have gained popularity in the past few years---due to increasingly better performance---with techniques such as Isomap~\cite{tenenbaum00}, LLE~\cite{roweis00}, or LAMP~\cite{joia11}; a few comparative review papers on general DR exist already, see the surveys \cite{van2009dimensionality} or \cite{Espadoto2019}.
This popularity has reached its peak after the publication of t-distributed Stochastic Neighbor Embedding (t-SNE)~\cite{vanDerMaaten:2008}. Through a series of complex transformations and fine-tuned optimization procedures (cf.~\autoref{sec:tsne}), t-SNE usually manages to create low-dimensional representations that capture complex patterns from the high-dimensional space very accurately, showing them as well-separated clusters of points. It has been used successfully in many different domains such as single-cell mass cytometry~\cite{hollt16}, natural language processing~\cite{johnson17}, and cancer analysis~\cite{amir13}.

t-SNE's inherent complexity, however, has also raised concerns regarding the trustworthiness of the results and the difficulty in interpreting them. Wattenberg et al.~\cite{wattenberg2016} demonstrated several important pitfalls of t-SNE, such as (i) the highly-complicated relationship between input parameters and visualization, (ii) the apparent irrelevance of the sizes (or density) of high-dimensional clusters, (iii) the disregard for the distance between clusters, (iv) the appearance of clusters even when the input is random, and (v) the difficulty in assessing and (vi) interpreting shapes. Although they also include advices and guidelines for using t-SNE effectively, the examples use simple and carefully-engineered artificial data sets, for which the original appearance is clear. Therefore, one question remains open: how to avoid such pitfalls with real-world high-dimensional data, possibly in the thousands of dimensions, when little or no previous knowledge is available?

Inspired by the work of Wattenberg et al.~\cite{wattenberg2016} and the existing visualization literature on interpreting and assessing DR methods~\cite{sacha17,nonato18}, 
we present t-viSNE, a tool designed to support the interactive exploration of t-SNE projections (an extension to our previous poster abstract~\cite{Chatzimparmpas2018}). In contrast to other, more general approaches, t-viSNE was designed with the specific problems related to the investigation of t-SNE projections in mind, bringing to light some of the hidden internal workings of the algorithm which, when visualized, may provide important insights about the high-dimensional data set under analysis. 
Our proposed solution is composed of a set of coordinated views that work together in order to fulfill four main goals: (G1) facilitate the \emph{choice of hyper-parameters} through visual exploration and the use of quality metrics; (G2) provide a quick \emph{overview of the accuracy} of the projection, to support the decision of either moving forward with the analysis or repeating the process of hyper-parameter exploration; (G3) provide the means to \emph{investigate quality} further, differentiating between the trustworthiness of different regions of the projection; and (G4) allow the \emph{interpretation} of different visible patterns of the projection in terms of the original data set's dimensions. 

The implemented views are a mix of adapted and improved classic techniques (e.g., our Shepard Heatmap and Adaptive Parallel Coordinates Plot (PCP)), new proposals (e.g., the Dimension Correlation view), and standard visual mappings with information that is usually hidden or lost after the projection is created (e.g., Density and Remaining Cost views). 
They were created in a careful design process that aimed to bring forward a selection of visualization techniques, combined and put together as a coherent whole in order to support---as much as possible---an accessible and usable analysis workflow with t-SNE. To the best of our knowledge, t-viSNE is the first interactive visualization tool designed with the goal of alleviating the specific shortcomings of t-SNE and supporting, at the same time and in a coherent and usable way, the assessment of quality and the interpretation of patterns in t-SNE projections. In summary, our contributions consist of 
\begin{itemize}
\item a selection of different views, interaction techniques, and visual mappings designed to support the interpretation and assessment of t-SNE projections;
\item their implementation in a carefully-designed system geared towards supporting analysts in overcoming well-documented difficulties of working with t-SNE; and
\item a discussion on the design and the outcomes of a user study that showed promising results.
\end{itemize}

\noindent Although our proposed solution is inspired by the work of Wattenberg et al.~\cite{wattenberg2016} and touches on most of the points raised by the authors, not all of them are fully covered by t-viSNE. More specifically, t-viSNE addresses points (ii), (iii), (v), and (vi) described previously, partially covers (i), and leaves point (iv) for future work, i.e., we only omit the investigation on how the formation of clusters might erroneously convey messages to the users even when the input is random. Thus, we intend this work to be a comprehensive proposal of possible solutions to the problem of opening t-SNE's black box, and to provide very important and relevant steps towards that final goal.

The rest of this paper is organized as follows. In the next two sections, we discuss literature that is related to visual, interactive assessment and interpretation of t-SNE projections as well as the necessary background information on how the t-SNE algorithm works.~\autoref{t-viSNE} presents our visualization approach including the various features of t-viSNE in the three categories: overview, quality, and dimensions. We then demonstrate the effectiveness of t-viSNE by describing two use cases with real data in~\autoref{case}. 
Thereafter in~\autoref{eval}, we discuss the usability and applicability of t-viSNE by reporting the results of a user study. 
~\autoref{sec:disc} discusses our design choices, limitations, and possible future work. Finally,~\autoref{conclus} concludes our paper.

%% file: 2.Related_Work.tex
A DR method is an algorithm that \emph{projects} a high-dimensional data set to a low-dimensional representation, preserving the structure of the original data as much as possible. 
Most of these algorithms have some (or many) hyper-parameters that may considerably affect their results, but setting them correctly is not a trivial task. In Subsection~\ref{sec:ParametersExpl}, we briefly describe techniques that try to solve this problem, and discuss the differences to our tool's functionality. The resulting projection is usually visualized with scatterplots, which support tasks such as finding groups of similar points, correlations, and outliers~\cite{nonato18}. However, a scatterplot is simply the first step in analyzing a high-dimensional data set through a projection: questions regarding the quality of the results (see Subsection~\ref{Quality}) and how to interpret them (see Subsection~\ref{Interpretation}) are pervasive in the literature on the subject. 
A few other tools have been proposed throughout the years that incorporate these techniques to deal with the problem of supporting the exploration of multidimensional data with DR. In Subsection~\ref{Other}, we discuss their goals and trade-offs, and compare them with t-viSNE.

Additionally, a summary of the tools discussed in this section and a feature comparison with t-viSNE is presented in Table~\ref{tab:featurescomp}. 
A tick indicates that the tool has the corresponding features/capabilities, while a tick in parentheses means the tool offers \emph{implicit} support (i.e., it could be done manually, in an ad hoc manner, but is not explicitly supported).
The table does not include works that do not contain a concrete visualization tool as their research contribution, as in Schreck et al.~\cite{schreck10}, for instance. Furthermore, we excluded the works which are not generalizable and focus on specific domain applications such as~\cite{sherkat18,endert12}.

\begin{table}[ht]
  \caption{Feature comparison of t-viSNE~\cite{tviSNECode} with other related tools from the literature. The last column indicates if the tool (T) and/or its source code (SC) are available online (last checked: January 15, 2020).} 
  \includegraphics[width=\linewidth]{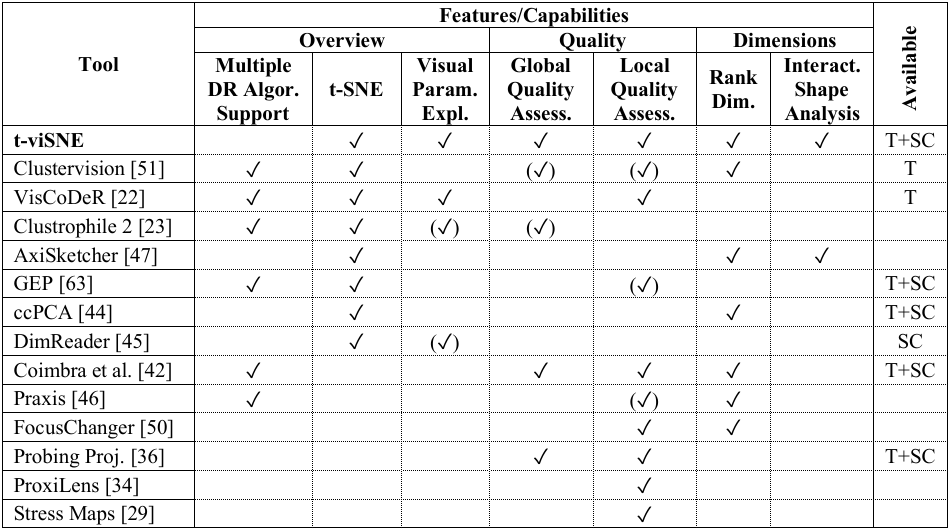}
  \label{tab:featurescomp}
  \vspace{-2em}
\end{table}

\subsection{Hyper-parameter Exploration} \label{sec:ParametersExpl}
VisCoDeR~\cite{Cutura18} supports the comparison between multiple projections generated by different DR techniques and parameter settings, similarly to our initial parameter exploration, using a scatterplot view with an on-top heatmap visualization for evaluating the quality of these projections. In contrast to t-viSNE, it does not support further exploratory visual analysis tasks after the layout is selected, such as optimizing the hyper-parameters for specific user selections. 
Clustrophile 2~\cite{cavallo19} contains a Clustering Tour feature to partially assist users in immediately exploring the space of potential clustering results by visualizing previous and current solution states, and providing choices of modalities by which the user can restrain how parameters are updated. These features help with the investigation of the quality of different clustering results (see the subsection below) in relation to the users' analytical tasks. However, t-viSNE supports the visual exploration of a predetermined space of solutions, which allows users to optimize sublocations and highlight these patterns with new projections.

\subsection{Quality Assessment} \label{Quality}

One way to obtain an indication of a projection's quality is to compute a single scalar value, equivalent to a final \emph{score}. Examples are Normalized Stress~\cite{joia11}, Trustworthiness and Continuity~\cite{venna06}, and Distance Consistency (DSC)~\cite{sips09}. More recently, ClustMe~\cite{Abbas19} was proposed as a perception-based measure that ranks scatterplots based on cluster-related patterns.
While this might be useful for quick overviews or automatic selection of projections, a single score fails to capture more intricate details, such as \emph{where} and \emph{why} a projection is good or bad~\cite{martins14}. In contrast, local measures such as the \emph{projection precision score} (pps)~\cite{schreck10} describe the quality for each individual point of the projection, which can then be visualized as an extra layer on top of the scatterplot itself. These measures usually focus on the preservation of neighborhoods~\cite{mokbel13,kohlhammer10,lee09} or distances~\cite{martins14,lespinats11,aupetit07}.
As an example, the \emph{set difference} from Martins et al.~\cite{martins15} uses the Jaccard set-distance between the two sets of neighbors of a point in low- and high-dimensional space in order to compute a measure of Neighborhood Preservation. We have chosen to adopt it in our work, in contrast to others, because of its intuitive interpretation, simple computation, and straightforward adaptation for displaying the preservation of neighborhoods of different scales.

Quality measures can also be used in interactive explorations to expose errors on demand and direct the user's further explorations~\cite{heulot13}. Liu et al.~\cite{liu14} use a combination of hierarchical clustering and different local measures to guide the user in manipulating the projection and testing hypotheses about the data. Probing Projections~\cite{stahnke16} simulates a correction of the points' positions according to a reference point, showing how a more accurate projection would look like. Fernstad et al.~\cite{fernstad13} use combinations of quality measures to determine the most interesting dimensions of the data and guide user exploration.
t-viSNE is similar to these works in its use of measures to guide the user's exploration, but we use measures and mappings that are either specific to t-SNE's algorithm or customized to be more useful in this scenario.
For more details about the assessment of quality in projections, we refer the reader to Nonato and Aupetit's recent survey~\cite{nonato18}.


\subsection{Interpretation of Projections} \label{Interpretation}
Some attempts to enrich scatterplots with automatically-derived statistical descriptions of patterns~\cite{silva15,kandogan12,chen10} have shown that static mappings may be useful in simple scenarios, but the complex relations between low- and high-dimensional space in non-linear projections cannot be well represented.
In such cases, interactive visual interfaces are necessary, as noted by Sacha et al.~\cite{sacha17} in their survey on interaction techniques for DR. 
Interactive solutions for specific domains such as text~\cite{sherkat18,endert12} and images~\cite{tan12,joia11} use inherent characteristics of the data in order to explain layouts, however, they are not easily generalizable to other domains.
In their tool, Coimbra et al.~\cite{coimbra16} support interactive exploration of 3-D projections using adapted biplots and different widgets for viewpoint selection. Our tool is similar to theirs from the perspective of providing a collection of interconnected views for projection exploration, but they focus on projection-agnostic 3-D scatterplots, and the widgets have different goals.
%
%
Probing Projections~\cite{stahnke16} is another such interactive system that supports both explaining and assessing projections, but limited to MDS~\cite{borg2003}. Groups of points can be compared in terms of the data set's dimensions, and a heatmap of the distribution of a selected dimension can be overlaid on the visualization, but there is no special prioritization of dimensions to deal with very high-dimensional data; the user must simply cycle through all of them in order to find the most relevant one.
Fujiwara et al.~\cite{fujiwara20} proposed the \emph{contrasting clusters in PCA} (ccPCA) method to find which dimensions contributed more to the formation of a selected cluster and why it differs from the rest of the dataset, based on information on separation and internal vs. external variability. We have similar goals, but approach them with different methods. For exploring clusters and selections in general, we use PCA to filter and order a local PCP plot; this could be easily adapted to use ccPCA instead as an underlying method for choosing which dimensions to filter and how to re-order the axes, without affecting the overall proposed analytical flow of the tool. On the other hand, ccPCA does not deal with the analysis of shapes, which we support with our proposed \emph{Dimension Correlation}.
%
%
Other recent approaches include DimReader~\cite{faust19}, where the authors create so-called \emph{generalized axes} for non-linear DR methods, but besides explaining a single dimension at a time, it is currently unclear how exactly it can be used in an interactive exploration scenario; and
Praxis~\cite{cavallo18}, with two methods---\emph{backward} and \emph{forward} projection---but it requires fast out-of-sample extensions which are not available for the original t-SNE. 

Most similarly to one of our proposed interactions (the \emph{Dimension Correlation}, Subsection~\ref{sec:dim_corr}), in AxiSketcher~\cite{kwon17} (and its prior version InterAxis~\cite{kim2016}) the user can draw a polyline in the scatterplot to identify a shape, which results in new non-linear high-dimensional axes to match the user's intentions. Since the resulting dimension contributions to the axes are not uniform, it is not possible to represent them using simple means such as bar charts. In our \emph{Dimension Correlation} tool, the user also draws a polyline to identify a shape, but our intention is exactly the opposite of AxiSketcher: we want to capture dimension contributions in an easy and accessible way. For this, we project low-dimensional points into the line (not high-dimensional ones, as in AxiSketcher), and we compute the dimension contributions in a different way, using Spearman's rank correlation. In summary, although there is a superficial similarity between the two techniques regarding how the user interacts with the scatterplot, their goals and their inner workings are quite different. Since t-viSNE adopts an approach of combining many different coordinated views, it is important for the \emph{Dimension Correlation} to maintain---as much as possible---the users' mental map of the projection, and to give simple and straightforward interpretations of the patterns they see.




\subsection{Comparison with Other Tools} \label{Other}

Other than the ones discussed so far, some interactive tools have been designed with either specific DR methods in mind, such as SIRIUS~\cite{dowling2019}, and FocusChanger~\cite{lai2018}, or for specific domains, such as Cytosplore~\cite{hollt16}. t-SNE can also be used to explore and judge different clustering partitions of the same data set, as in Clustervision~\cite{kwon2018}. 


SIRIUS~\cite{dowling2019} focuses on the concurrent exploration of similarity relationships between instances and between dimensions, analyzing their relationship and providing interaction techniques via a dual visualization approach, with two coordinated side-by-side scatterplots. In t-viSNE, we focus on bringing forward hidden information about the DR algorithm that is usually lost, with all the interactions occurring in a single main scatterplot view (and some additional auxiliary views). One of our goals is to also support the user in testing the quality of the algorithm to increase its trustworthiness, a task that is not supported by SIRIUS.

FocusChanger~\cite{lai2018} empowers users to perform local analyses by setting Points of Interest (POIs) in a linear projection, which is then updated to enhance the representation of the selected POIs. When hovering over specific points, the information of true neighborhood of other points is mapped to the saturation of the color. This allows for a simple mechanism of quality assessment, but hurts the possibility of using color for other mappings and requires pointwise interaction. The used projections are linear and, thus, potentially not as representative and useful as t-SNE. Similar to Andromeda, it relies on the possibility of quickly updating them, which might not be currently feasible with t-SNE.

Cytosplore~\cite{hollt16} is an example of tools that use t-SNE for visual data exploration within a specific domain: single-cell analysis with mass cytometry data. Apart from showing a t-SNE projection of the data, Cytosplore is also supported by a domain-specific clustering technique which serves as the base for the rest of the provided visualizations, but is not generalizable to other domains. 
%
%

Clustervision~\cite{kwon2018} is a visualization tool used to test multiple batches of a varying number of clusters and allows the users to pick the best partitioning according to their task. Then, the dimensions are ordered according to a cluster separation importance ranking. As a result, the interpretation and assessment of the final results are intrinsically tied to the choice of clustering algorithm, which is an external technique that is (in general) not related to the DR itself. Thus, the quality of the results is tied to the quality of the chosen clustering algorithm. With t-viSNE it is also possible to explore the results of a clustering technique by, for example, mapping them to labels, then using the labels as regions of interest during the interactive exploration of the data. However, the labels do not influence the results of t-viSNE, whether they exist or not, since we did not intend to tie the quality of our results to other external (and independent) techniques.

%% file: 3.t-SNE.tex
All the details about t-SNE's algorithm have been exhaustively described since its first publication (see, e.g., \cite{VanDerMaaten14,pezzotti17,chan18}). Here, we give a quick overview of the general steps of the algorithm and focus mostly on the specific details that are important for understanding the features of our tool.

The input to t-SNE is an $n \times N$ data matrix $X$, composed of a set of $n$ instances $x_i$ (rows) in $N$ dimensions (columns). As the first step, pairwise distances between instances are transformed into probability distributions that represent neighborhoods in the following way. For every pair of instances $(x_i, x_j)$ with $1 \leq i,j \leq n$, a probability $p_{ij}$ is computed as
\begin{equation} \label{eq:pji}
	p_{ij} = \frac{p_{j|i}+p_{i|j}}{2n}, \quad p_{j|i} = \frac{exp(-\|x_i-x_j\|^2/2\sigma_i^2)}{\sum_{k \neq l}{exp(-\|x_k-x_l\|^2/2\sigma_i^2)}}.
\end{equation}

\autoref{eq:pji} can be interpreted as \textit{the probability that two instances $x_i$ and $x_j$ would pick each other as close neighbors}. It is roughly equivalent to centering a multivariate Gaussian around $x_i$ and setting $p_{j|i}$ to the Gaussian-transformed value of its distance to $x_j$, then the same but centered on $x_j$, and finally combining both. That translates to high probabilities for near neighbors and very small probabilities for farther ones. One of the most important things to notice from Equation~\ref{eq:pji}, however, is that the variance of the Gaussian, i.e., $\sigma_i$, is different for each $x_i$: that means that the \textit{bandwidth} of the Gaussian changes for each high-dimensional instance, in order to capture the variations in density for different high-dimensional neighborhoods. This value is found iteratively by trial-and-error, using binary search, until a user-defined \textit{perplexity} is reached, with $Perp(i)=2^{H(i)}$ and $ H(i)=-\sum_j{p_{j|i} log_{2} p_{j|i}}$.

Considering $P$ as the joint distribution including all pairwise probabilities computed according to Equation~\ref{eq:pji}, the goal of t-SNE is, then, to find another probability distribution $Q$ that faithfully represents $P$ in low-dimensional spaces, usually in 2-D or 3-D (to allow for their straightforward visualization). Each pair of low-dimensional points $(y_i, y_j)$ is also modeled as a probability, now called $q_{ij}$, as
\begin{equation}
	q_{ij}=\frac{(1+\|y_i-y_j\|^2)^{-1}}{\sum_{k \neq l}{(1+\|y_k-y_l\|^2)^{-1}}}.
\end{equation}

Instead of using Gaussians again, a Student's t-distribution with one degree of freedom is used for $Q$. Notice that, as opposed to $P$, the distribution of $Q$ is not parameterized with a variable neighborhood density (i.e., there is no $\sigma_i$). This means that, potentially, neighborhoods with very different densities in the original high-dimensional space may be mapped into areas of equivalent size in the low-dimensional representation.
 
The search for a $Q$ that faithfully represents $P$ in a low-dimensional space is done by optimizing a cost function ($C$) given by the Kullback-Leibler ($KL$) divergence between the two distributions,
\begin{equation}
	C=KL(P\|Q)=\sum_i{KL(P_i\|Q_i)}, \quad KL(P_i\|Q_i) = \sum_j{p_{ij}log\frac{p_{ij}}{q_{ij}}},
\end{equation}
which is performed with gradient descent for a user-specified number of iterations. In each iteration, every point $y_i$ is adjusted towards the direction of the largest decrease in its associated cost $KL(P_i\|Q_i)$, i.e., the Kullback-Leibler Divergence ($KLD$) between the low-dimensional neighborhood of $y_i$ and the high-dimensional neighborhood of $x_i$. Computing this cost involves comparing $y_i$ with all other points, which results in a complexity of $O(N^2)$. The final remaining cost $C$ after the optimization is, then, a sum of all the remaining costs $KL(P_i\|Q_i)$.


It is important to notice that the original t-SNE algorithm has been updated and accelerated in many different ways throughout the years, most famously by the original author~\cite{VanDerMaaten14}, but also by other researchers using techniques such as approximations~\cite{pezzotti17} and parallel computing~\cite{chan18}. These newer versions give mostly accurate results, but are not completely exact. Please refer to Subsection~\ref{sec:limit} for a discussion on why we chose to use the Barnes–Hut t-SNE algorithm in this paper~\cite{VanDerMaaten14}.


%% file: 4.System_Overview.tex
\begin{figure*}[ht]
  \centering
  \includegraphics[width=\linewidth]{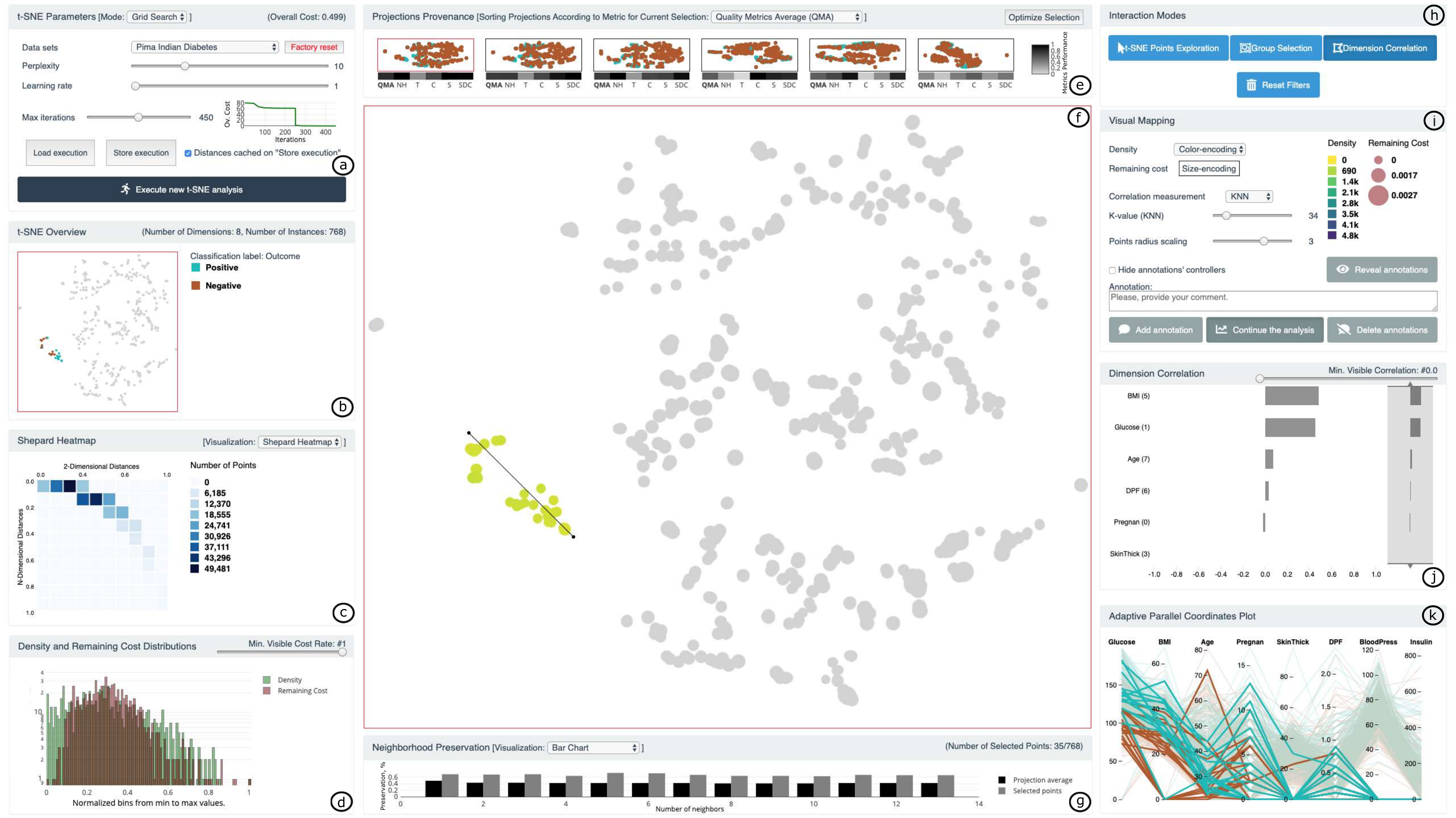}
  \caption{Visual inspection of t-SNE results with t-viSNE: (a) a panel for uploading data sets, choosing between two execution modes (grid search or a single set of parameters), and storing new (or loading previous) executions; (b) overview of the results with data-specific labels encoded with categorical colors; (c) the Shepard Heatmap of all pairwise distances; (d) the histogram with the Density and Remaining Cost distributions; (e) list of available projections, ranked by quality; (f) the main scatterplot view representing the Density of neighborhoods in the original high-dimensional space and the Remaining Cost of each point; (g) the Neighborhood Preservation bar chart/line plot; (h) control elements for the different interaction modes of the tool; (i) the visual mapping panel with a variety of options for the users such as an annotation tool for saving notes for multi-session analyses; (j) the Dimension Correlation bar chart visualizing the correlations between the data dimensions; and (k) the Adaptive PCP plot representing the most important dimensions.}
  \label{fig:teaser}
\end{figure*}

Most of the related works described in~\autoref{sec:relwork} deal with the problem of assessing and interpreting DR in general, and aim to be applicable to a wide range of different scenarios, providing solutions that overlook the specific shortcomings of each DR method. While this approach has its merits, a gap remains regarding the treatment of method-specific problems that might lead to more directly-applicable results. However, very few single DR methods have enough widespread acceptance to warrant customized treatments (with the exception of PCA and MDS, for example). Nowadays, arguably, the situation has changed: t-SNE is almost a standard DR method for both analysts and researchers. Due to this, it is our understanding that a set of methods that is specifically designed to meet t-SNE's shortcomings deserves its place among the current body of work in the interpretation and assessment of DR methods, and its potentials are large enough to deserve their own treatment.

In this section we describe t-viSNE, a web-based system that implements an assortment of views and interaction tools that bring to light many facets of a t-SNE projection which are usually hidden behind its black box.
We aim to enhance the trust into and interpretability of t-SNE through visualization and exploration of the model, the data, and the hyper-parameters. An overall picture of the interface is shown in~\autoref{fig:teaser}, and each of its different views is described below, divided into our four design goals: \emph{Hyper-parameter Exploration (G1)}, \emph{Overview} (G2), \emph{Quality} (G3), and \emph{Dimensions} (G4). Further discussions on the design choices behind some of the views can be found in Subsection~\ref{sec:design}.

\subsection{Goal 1: Hyper-parameter Exploration} \label{sec:parameters}

\begin{figure}[tb]
  \centering
  \includegraphics[width=\linewidth]{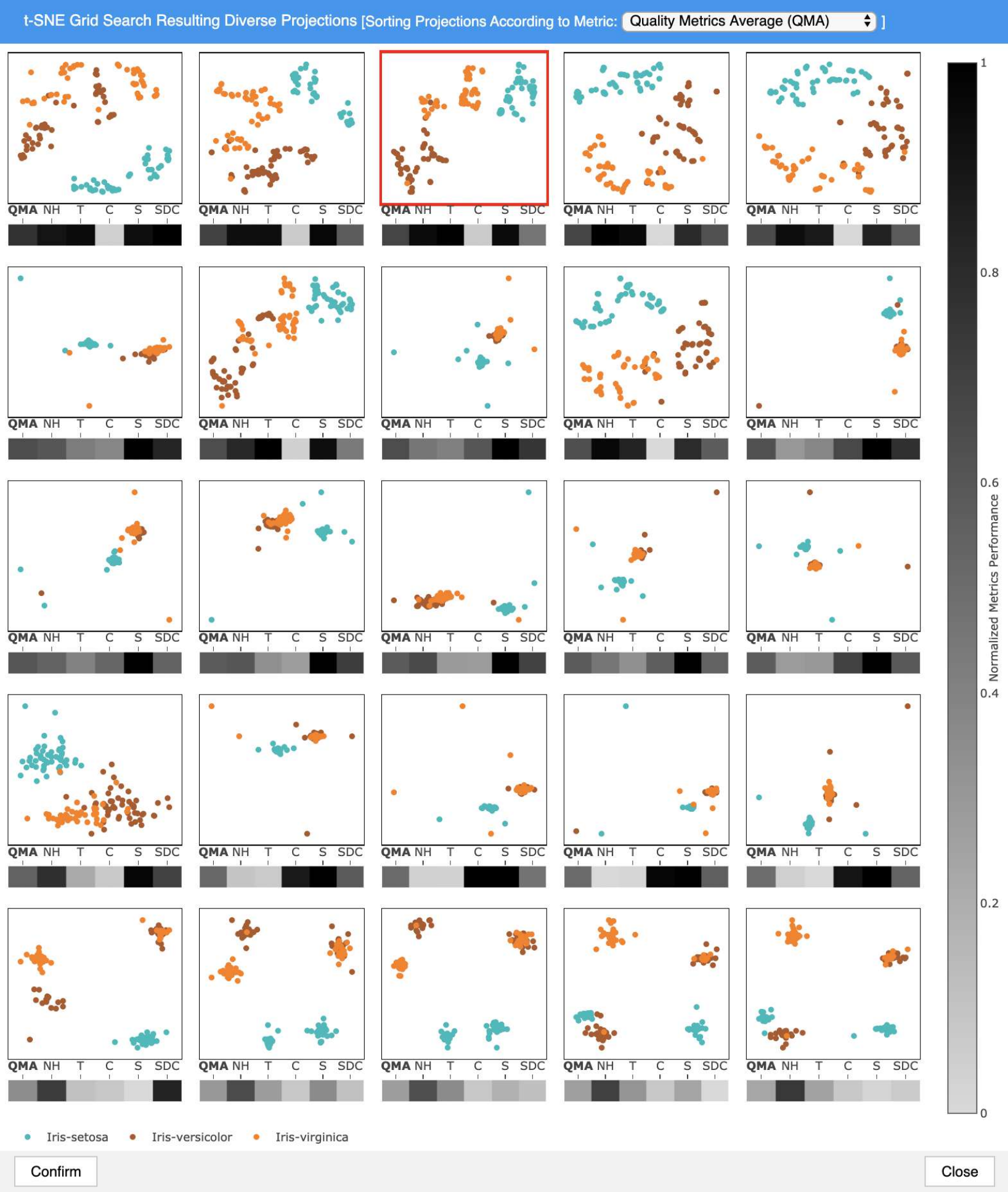}
  \caption{Hyper-parameter exploration (presented in a dialog at the beginning of an analytical session), with 25 representative projections from a pool of 500 alternatives obtained through a grid search. Five quality metrics, plus their Quality Metrics Average (QMA), are also displayed to support the visual analysis. The thumbnails are sorted according to the QMA and ordered row-wise from top to bottom. The currently-selected projection is indicated by a red box (top row, third column).}
  \label{fig:parameters}
\end{figure}

Significantly-different t-SNE projections can be generated from the same data set, due to its well-known sensitivity to hyper-parameter settings~\cite{wattenberg2016}. We propose to support users in finding a good t-SNE projection for their data by using visual exploration, as follows. A \emph{Grid Search} mode (\autoref{fig:teaser}(a)) initiates a systematic parameter search that computes 500 projections by varying the parameters \emph{perplexity}, \emph{learning rate}, and \emph{max iterations}.
From this pool of 500 projections, 25 representative examples are singled out and shown to the user---in a matrix of thumbnails depicted in Figure~\ref{fig:parameters}---as suggestions of possible projections of the data. In order to choose the representatives, we partition the pool of 500 projections into 25 clusters (with K-Medoids~\cite{Kaufman1987Clustering}), using Procrustes distance~\cite{Duta2015Procrustes} as the dissimilarity measure. The medoids of the 25 resulting clusters are used as representatives. This whole process is transparent to the user and happens in the backend; only the representatives are shown. 
We give extra support to the user by providing the results of 5 quality measures for each representative projection: neighborhood hit (NH), trustworthiness (T), continuity (C), normalized stress (S), and Shepard diagram correlation (SDC), accompanied by the quality metrics average (QMA). They are shown as a grayscale heatmap under each cell of the thumbnail matrix (Figure~\ref{fig:parameters}). For more details on the quality measures, please refer to~\cite{Espadoto2019}.
It is important to clarify, however, that these quality measures are offered only as a support for the visual analysis. The main goal here is not to show the 25 \emph{best} projections, but the most \emph{diverse} ones; it is then the task of users---through visual exploration and by matching their own personal preferences---to choose the one that looks more promising.

After choosing a projection, users will proceed with the visual analysis using all the functionalities described in the next sections. However, the hyper-parameter exploration does not necessarily stop here. The top 6 representatives (according to a user-selected quality measure) are still shown at the top of the main view (Figure~\ref{fig:teaser}(e)), and the projection can be switched at any time if the user is not satisfied with the initial choice. We also provide the mechanism for a \emph{selection-based ranking} of the representatives. During the exploration of the projection, if the user finds a certain pattern of interest (i.e., cluster, shape, etc.), one possible question might be whether this specific pattern is better visible or better represented in another projection. After selecting these points, the list of top representatives can be ranked again to contain the projections with the best quality regarding the selection (as opposed to the best global quality, which is the default). The way this ``selection-based quality'' is computed is by adapting the global quality measures we used, taking advantage of the fact that they all work by aggregating a measure-specific quality computation over all the points of the projection. In the case of the selection-based quality, we aggregate only over the selected points to reach the final value of the quality measure, which is then used to re-rank the representatives.

\subsection{Goal 2: Overview} \label{sec:overview}

The main view of the tool (\autoref{fig:teaser}(f)) presents the t-SNE results as an interactive scatterplot, with specific mappings on the points' colors and sizes (see Subsection~\ref{sec:vis_map} for details). There are four \emph{Interaction Modes} (\autoref{fig:teaser}(h)) for this view, as described next. The first (and default) mode---\emph{t-SNE Points Exploration}---activates panning, zooming, and hovering, supporting the user to focus on individual patterns of the projection, and to investigate specific points' dimensions. The second mode---\emph{Group Selection}---provides a lasso selection tool that triggers updates in other views, such as the Neighborhood Preservation view (Subsection~\ref{sec:neigh_pres}) and the Adaptive PCP (Subsection~\ref{sec:lasso_pcp}). 
The third option---\emph{Dimension Correlation}---provides a tool for the user to check the hypothesis that a visual pattern, as observed, is strongly correlated to a pattern in the high-dimensional space (Subsection~\ref{sec:dim_corr}). The final mode---\emph{Reset Filters}---removes every filter applied with the previously-described interaction modes. 

To complement the main view, the \emph{Overview} (\autoref{fig:teaser}(b)) shows the static t-SNE projection and serves as a contextual anchor that is independent of the interactions and/or filters applied to the main view. Data-specific labels (when those exist) are shown using a categorical colormap, along with simple statistics about the data set. 

\subsection{Goal 3: Quality} \label{sec:vis_map}\label{sec:neigh_pres}

Before the users move on with a more detailed interpretation of the patterns that are visible in the scatterplot resulting from t-SNE's projection, it is important that they \emph{trust} what they see. We approach the investigation of quality both globally, with simplified and aggregated views for the entire projection, and locally, so that the users can check if specific visible patterns are indeed present in the original space of the data set.

\mypar{Shepard Heatmap}
A Shepard Diagram~\cite{Leeuw2015} is a common way of assessing the accuracy of a visualization produced by a projection method. It consists of a scatterplot where each point represents a pair of instances from the data set. The value of the $y$-axis indicates their distance in the $N$-dimensional ($N$-D) space, and the $x$-axis their $2$-D distance. Both axes are scaled between $0.0$ (minimum distance) and $1.0$ (maximum distance), with the origin located on the top-left. 
For large data sets, however, such a scatterplot may become hard to read due to the very large number of points (in the order of $n^2$). To avoid this clutter problem and increase the readability of the Shepard Diagram for large data sets, we propose the Shepard Heatmap (\autoref{fig:teaser}(c)), which is an aggregated version of the Shepard Diagram, with the information of the number of points in each cell mapped to a single-hue colormap.

The main goal of the Shepard Heatmap is to offer a broad, simplified overview of the accuracy of the projection in terms of distance preservation: cells close to the main diagonal of the heatmap indicate that the respective pairs of instances have been represented in the $2$-D space with distances that are \emph{comparable} to their original $N$-D distances. Although it is well-known that t-SNE's goal is not to preserve distances~\cite{vanDerMaaten:2008}, but neighborhoods, the Shepard Heatmap still provides useful information to the analyst: if the cell values are closer to the $y$-axis than to the $x$-axis, then a large part of the data has been \emph{compressed}, i.e., a diverse range of distances from the $N$-D space have been represented with small distances in $2$-D. The opposite scenario (cell values being closer to $x$ than $y$) indicates that a small range of $N$-D distances have been \emph{spread} in a wide spectrum of distances in the $2$-D visualization.

\mypar{Visual Mapping}
The \emph{Visual Mapping} panel (\autoref{fig:teaser}(i)) includes controls for mapping \emph{Density} ($1/\sigma_i$) and \emph{Remaining Cost} ($KLD(P_i||Q_i)$) of each point to either color or size in the main view. These correspond to information extracted from the t-SNE algorithm itself, which would otherwise be hidden from the analyst. Their inspection, however, may prove fruitful, as we describe next.

\begin{figure}[tb]
  \centering
  \includegraphics[width=\linewidth]{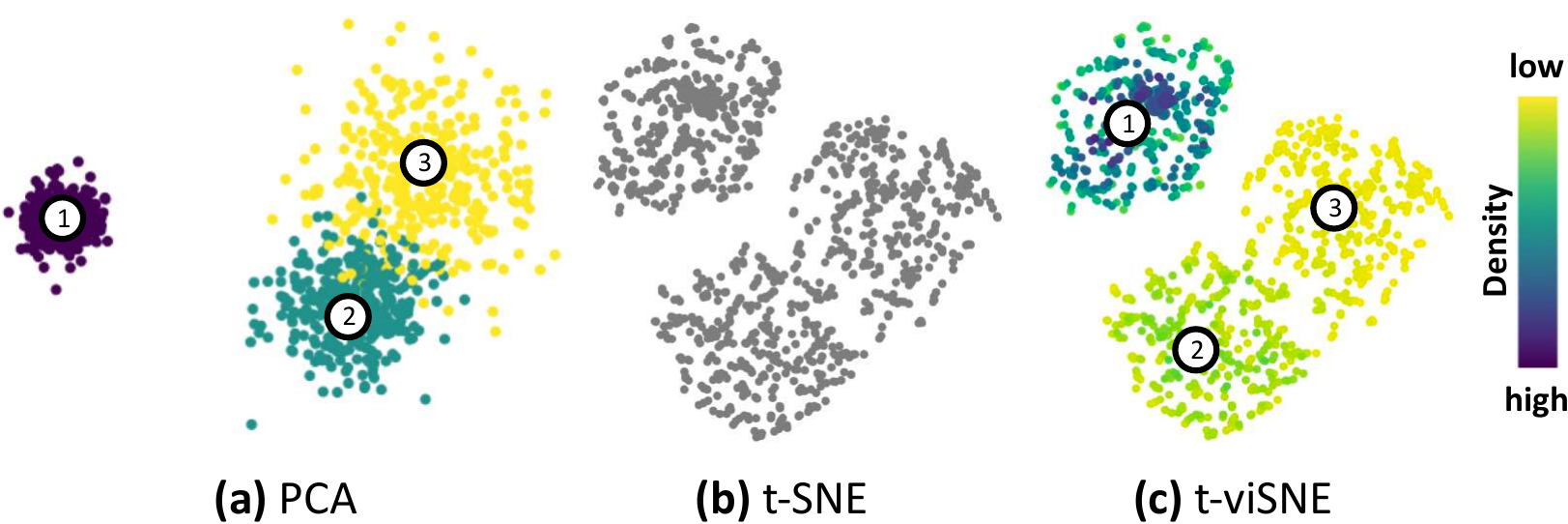}
  \caption{The importance of the visual mapping of \emph{Density}, using three $5$-D Gaussian clusters with varying standard deviations and slight overlap. (a) A simple linear projection using PCA shows the clusters' varying density. (b) A t-SNE projection shows all clusters with roughly the same size. (c) t-viSNE accurately shows the densities of the clusters (color-encoded) and helps us identify, for example, that clusters 2 and 3 are separate.}
  \label{fig:blob_dens}
\end{figure}

As we discussed in~\autoref{sec:tsne}, when t-SNE models the \mbox{$N$-D} space as probability distributions, each instance is assigned a different $\sigma_i$ that represents the \emph{Density} of that instance's original neighborhood. However, during the projection to the low-dimensional representation ($2$- or $3$-D), this information is usually lost, and neighborhoods with different densities appear to be very similar. Consider the simple example from~\autoref{fig:blob_dens}, where three $5$-D Gaussian clusters (with varying densities) are projected into $2$-D using PCA and t-SNE. The linear projection of PCA shows quite clearly that the clusters have different densities. The t-SNE projection, on the other hand, shows three clusters that are basically identical. We propose to recover this lost density information by extracting the values of $\sigma_i$ from the t-SNE process and mapping them on top of the points (using a sequential colormap, by default). The actual mapping is done with $\sigma_i^{-2}$, so that higher densities (lower values of $\sigma_i$) are mapped to higher values. As an example of the practical consequences of such a mapping, the visualization of the different density profiles of clusters 2 and 3 in t-viSNE (\autoref{fig:blob_dens}(c)) helps to identify that they are separate clusters and not a single large one, which could have been an erroneous insight in case no extra information was available (as in \autoref{fig:blob_dens}(b)).


\begin{figure}[htb]  
  \centering
  \includegraphics[width=\linewidth]{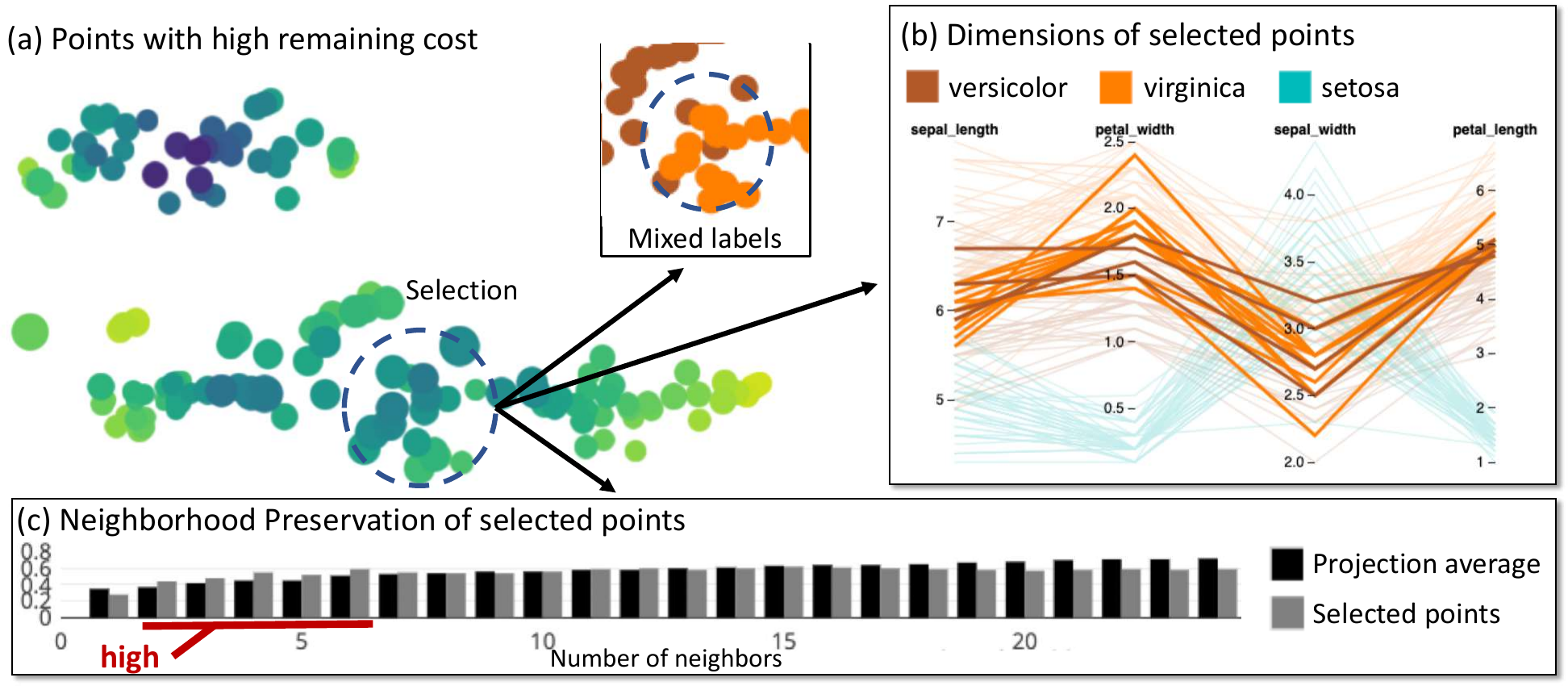}
  \caption{Investigation of a group of points from the well-known Iris data set~\cite{Dua2017Machine}. (a) The points' sizes indicate that a region in-between the species \emph{versicolor} and \emph{virginica} has the highest Remaining Cost. (b) The points have similar dimension values, but are classified as different species. (c) Neighborhood Preservation starts high (for close neighbors), but steadily decreases.}
  \label{fig:cost_pcp}
\end{figure}

The second option of the Visual Mapping panel, the \emph{Remaining Cost}, indicates (in the points' sizes, by default) the final value of $KLD(P_i\|Q_i)$ for each instance $x_i$, i.e., the remaining cost for each instance after the last iteration of t-SNE's optimization procedure (see~\autoref{sec:tsne}). It is common for the information of the overall remaining cost ($KLD(P\|Q)$) to be used as a direct judgment of the projection's quality. However, this perspective is limited, because a low overall remaining cost does not mean that the entire projection is equally good (and vice-versa for a high overall remaining cost). This is related to the idea of \emph{local} quality measures that has been motivated and explored in different previous work (see~\autoref{sec:relwork}), and shares the potential advantages of these measures. Hence, it allows the analyst to investigate which points (or groups of points) were harder to optimize according to t-SNE's cost function and, thus, affects the perception of the local trustworthiness of different areas of the projection. 
A simple example is shown in~\autoref{fig:cost_pcp}, using the well-known Iris data set~\cite{Dua2017Machine}. A group of points with high remaining cost can be found in the middle of the largest cluster in \autoref{fig:cost_pcp}(a). This cluster is, actually, a mix of two different species (\emph{versicolor} and \emph{virginica}), and the points with high remaining cost belong to the area where the two species are mixed, indicating instances where the $2$-D mapping might not be as straightforward as the rest. The dimensions of the selected points are highlighted in~\autoref{fig:cost_pcp}(b) using a PCP (see Subsection~\ref{sec:lasso_pcp}), confirming that these points are indeed characterized by dimension values that are relatively common to both species, which makes them harder to separate into isolated clusters.


\mypar{Neighborhood Preservation} 
Since the proposal of non-linear DR methods, 
the idea of prioritizing the preservation of close neighborhoods instead of pairwise distances in projections has been accepted as a positive trade-off, especially in visualization scenarios. The t-SNE algorithm also follows this idea: by transforming the pairwise distances into probability distributions using Gaussians (cf.~\autoref{sec:tsne}), it aims to preserve only the closest neighbors of each point, effectively ignoring farther ones. Due to this, the ability to investigate the extent to which such neighborhoods are preserved is one important piece of the puzzle that forms a full assessment of the accuracy of a t-SNE projection.

We present a \emph{Neighborhood Preservation} plot (\autoref{fig:teaser}(g)) that shows an overview of the preservation of neighborhoods of different sizes ($k$) in both the entire projection and the current selection, based on the Jaccard distance between sets:
\begin{equation}
  NP_k = \sum_i{\frac{1}{n} \cdot \frac{\nu_k^2(i) \cap \nu_k^N(i)}{\nu_k^2(i) \cup \nu_k^N(i)}},
\end{equation}
where $\nu_k^2(i)$ is the $k$-neighborhood of instance $i$ in $2$-D, $\nu_k^N(i)$ is the $k$-neighborhood of instance $i$ in $N$-D, and $n$ is the number of selected points (or the size of the data set, if nothing is selected). For each value of $k$, $NP_k$ yields the \emph{average preservation of neighborhoods} of up to $k$ points, centered at the $n$ selected points (or for the entire projection, if nothing is selected). This is an aggregated and interactive adaptation of ideas introduced by, for example, Joia et al.~\cite{joia11} and Martins et al.~\cite{martins15}.
The default visualization for the Neighborhood Preservation is a bar chart (as described below), but users have two more options to visualize the same information using line plots (see Subsection~\ref{sec:design} for a discussion and comparison).

The black bars are always fixed, showing the average preservation for all points of the projection. For example, in~\autoref{fig:cost_pcp}(c), the relatively tall black bars starting from the point $k = 20$ mean that, on average, neighborhoods of 20 points or more are well preserved. The same rationale applies to the gray-colored bars. However, their values change in connection with the lasso selection, so that they always show an up-to-date view of the Neighborhood Preservation centered at the selected group of points. This allows the analyst to compare them to the rest of the projection to get a \emph{relative} assessment, which is important since there are no absolute rules as to how much preservation is good or bad; such insights depend on the scale of the data set and of each high-dimensional pattern. In~\autoref{fig:cost_pcp}(c), for example, the tall gray-colored bars around $k = 4$ mean that, on average, neighborhoods of around 4 points are well preserved \emph{for the selected points}. This is in contrast to the overall preservation, which starts low and grows slowly with $k$. Since the selected points are positioned at the border between the two species clusters, they have very close near neighbors (i.e., points which are located in-between species), but as the value of $k$ grows, their neighborhoods become more mixed and, thus, less well-preserved.

\subsection{Goal 4: Dimensions} \label{sec:dimensions} \label{sec:lasso_pcp} \label{sec:dim_corr}

Having established trust in the visualization, the users then proceed to identify and investigate the visible patterns from the projected data. One of the most common analytical tasks in any DR-based workflow is, for example, to identify clusters of similar points~\cite{nonato18}, with the goal of detecting patterns in the organization of the data in the high-dimensional space. Irregularly-shaped clusters are also of interest~\cite{wattenberg2016}, which suggests that the points' organization along a non-linear multidimensional axis might be relevant. The problem of explaining the reasons why those clusters are formed is tackled by a number of t-viSNE views that are described next.

\mypar{Adaptive Parallel Coordinates Plot} 
Our first proposal to support the task of interpreting patterns in a t-SNE projection is an \emph{Adaptive} PCP~\cite{inselberg90}, as shown in~\autoref{fig:teaser}(k). It highlights the dimensions of the points selected with the lasso tool, using a maximum of 8 axes at any time, to avoid clutter. The shown axes (and their order) are, however, not fixed, as is the usual case. Instead, they are adapted to the selection in the following way. First, a Principal Component Analysis (PCA)~\cite{jolliffe16} is performed using \emph{only the selected points}, but with \emph{all} dimensions. That yields two results: (1) a set of eigenvectors that represent a new base that best explains the variance of the selected points, and (2) a set of eigenvalues that represent how much variance is explained by each eigenvector. Simulating a reduction of the dimensions of the selected points to $1$-Dimensional space using PCA, we pick the eigenvector with the largest eigenvalue, i.e., the most representative one. This $N$-D vector can be seen as sequence $w$ of $N$ weights, one per original dimension, where the value of $w_j$ indicates the importance of dimension $j$ in explaining the variance of the user-selected subset of the data. Finally, we sort $w$ in descending order, then pick the dimensions that correspond to the first (up to) 8 values of the sorted $w$. These are the (up to) 8 dimensions shown in the PCP axes, in the same descending order (from left to right).

Apart from the adaptive filtering and re-ordering of the axes, we maintained a rather standard visual presentation of the PCP plot, to make sure it is as easy and natural as possible for users to inspect it. The colors reflect the labels of the data with the same colors as in the \emph{overview} (Subsection~\ref{sec:overview}), when available, and the rest of the instances of the data---which are not selected---are shown with high transparency. Each axis maps the entire range of each dimension, from bottom to top. A simple example is given in~\autoref{fig:cost_pcp}(b), where we can see that the dimensions of the selected points roughly appear at the intersection between two species, versicolor (brown) and virginica (orange). 

\mypar{Dimension Correlation} 
Supporting the interpretation of clusters is definitely one important step towards interpreting t-SNE, but it does not cover the entire picture. As it has been noted by Wattenberg et al.~\cite{wattenberg2016}, t-SNE commonly generates visual patterns with different shapes, which may or may not faithfully represent the actual shapes of the original high-dimensional patterns. It is natural to expect that the user, upon seeing an oddly-shaped pattern, will come up with different hypotheses about why that shape exists, or at least will be curious to try to understand what exactly caused such a shape to appear.

\begin{figure}[tb]  
  \centering
  \includegraphics[width=\linewidth]{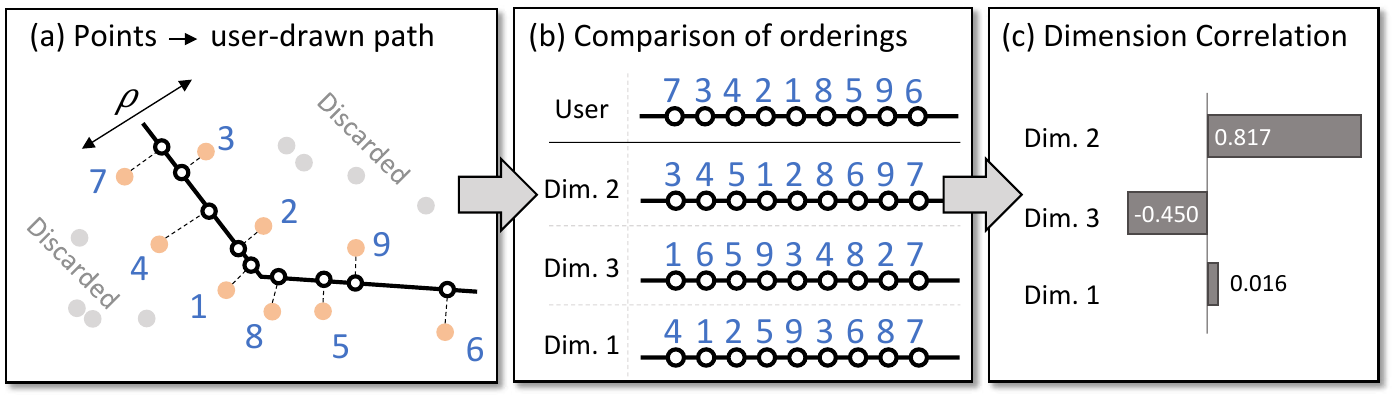}
  \caption{The \emph{Dimension Correlation} tool. (a) Nearby points are projected to a user-drawn path, creating a user-induced ordering. Here 7, 3, 4, and so on are data instance IDs. (b) The user-induced ordering is compared to dimension-specific orderings using a correlation measure. (c) Results are shown in the lengths of bars, ordered by the absolute value of the correlation (with highest on top). Note that if the same polyline is drawn by the user in the opposite direction over a pattern, then the signs of the correlations change but not their magnitude.}
  \label{fig:dim_corr}
\end{figure}

We propose the \emph{Dimension Correlation} tool, a novel interactive tool to explore and interpret such shapes in a t-SNE projection. It is triggered by a user interaction that consists of drawing a polyline with the mouse (i.e., a sequence of connected line segments), following the shape of the pattern detected by the user. After the polyline is finished, all points within a user-defined range $\rho$ of the polyline are selected and ``projected'' onto the polyline, in the following way (cf.~\autoref{fig:dim_corr}): (1) we find the minimum distance $d^p_i$ between each point $i$ in the scatterplot to the polyline $p$, defined as the minimum distance from $i$ to \emph{any} segment of $p$; (2) every point $i$ such that $d^p_i > \rho$ is discarded; and (3) for the remaining points, we find the point $p_i$ that is the projection of $i$ into $p$, i.e., the projection of $i$ into the segment of $p$ that is closest to $i$. 

If we ignore the actual distances between the points $p_i$ obtained along the polyline, a user-defined \emph{ordering} can be induced (or extracted) for the points $i$ that were not discarded during the process (cf.~\autoref{fig:dim_corr}(b)). This is one possible way of modeling---in a simple and unambiguous way---the shape of the visual pattern perceived by the user. Based on this ordering, we can then investigate which dimensions are more correlated to the pattern, i.e., are more relevant to explain its significance. For that, we first generate a set of dimension-specific orderings for the same points $i$ that were projected onto the polyline, using the values of these points along each dimension for the ordering (cf.~\autoref{fig:dim_corr}(b)). For example, in a data set $X$, for the dimension-specific ordering of dimension $j$, the values $X_{i,j}$ will be used (for the selected points $i$). The \emph{relevance} of each dimension is then defined as the absolute value of the correlation between its dimension-specific ordering and the user-defined ordering of the points $i$, which is equivalent to the Spearman's rank correlation coefficient~\cite{corder14}. We use the absolute value here, because the fact that the correlation is positive or negative is not critical. A strong negative correlation simply means that the pattern goes in the opposite direction of the one used when drawing the polyline.



The results (i.e., relevances of each dimension) are finally shown in an interactive horizontal bar chart (\autoref{fig:teaser}(j)), where the dimensions are sorted from top to bottom according to relevance (with the most relevant on the top). While the relevance is computed using the \emph{absolute} value of the correlation, we decided to show the \emph{original} value in the bars (including negative correlations to the left of the central axis) to avoid possibly misleading the analyst. This is illustrated in~\autoref{fig:dim_corr}(c). The final component of the Dimension Correlation tool is the ability to explore the different dimensions by clicking on the bars, which will change the colormap of the main view to reflect the values of the points for that specific dimension.

It is important to notice that the goal of the Dimension Correlation tool is not to dictate exactly which are the dimensions that cause the formation of a shape in a t-SNE projection. We propose a way to suggest the most interesting dimensions according to a detected visual pattern, in order to help analysts to prioritize the dimensions they will investigate further. Mapping the values of specific dimensions on top of the points of the scatterplot (usually with colors) is a common way to try to find relationships between dimensions and patterns during the exploration of DR projections. Without any support, it is also usually a cumbersome activity for high-dimensional data sets, requiring analysts to cycle through a large number of dimensions. Our intention with the Dimension Correlation tool is to work towards closing this gap.


\begin{figure*}[tb]
  \centering
  \includegraphics[width=\linewidth]{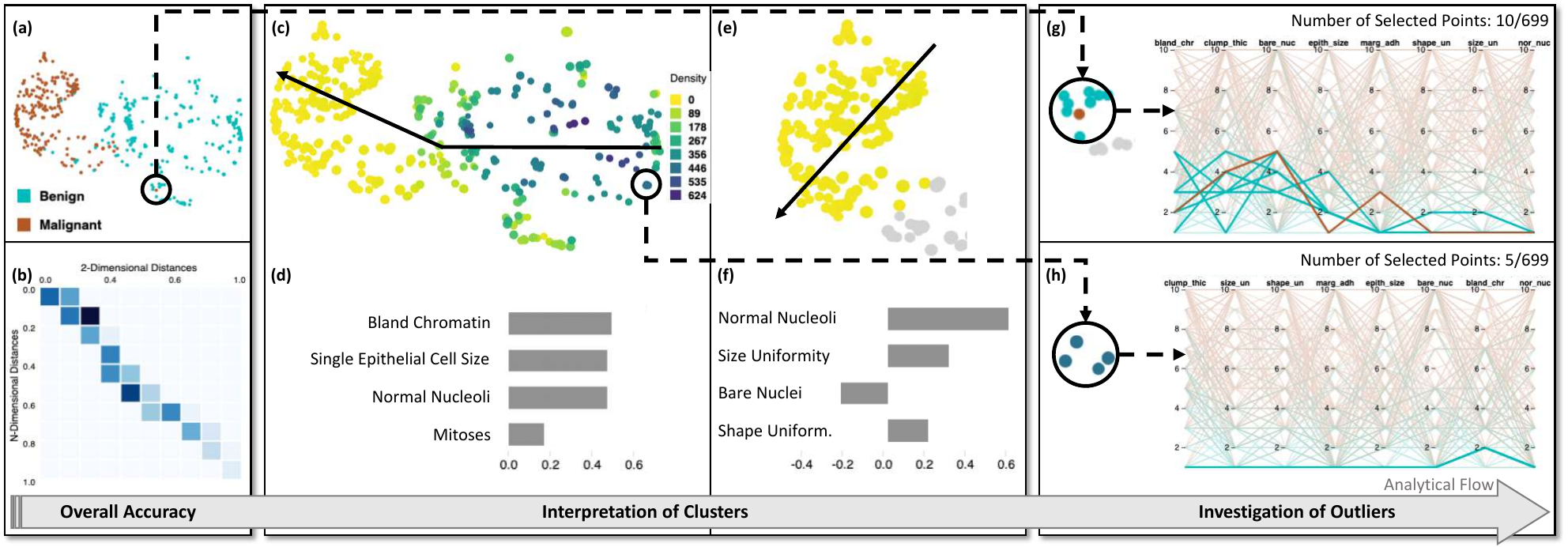}
  \caption{Usage scenario based on the Breast Cancer Wisconsin data set. The Overview (a) and the Shepard Heatmap (b) indicate that the \emph{overall accuracy} is good. The high density of \emph{benign} cases (c) seems to indicate that their high-dimensional profile is clearer and less diverse than \emph{malignant} cases, which are more sparse. Different combinations of dimensions are correlated with patterns between clusters (c, d) and inside clusters (e, f), which affects the \emph{interpretation of clusters}. The \emph{investigation of outliers} leads to identifying points that are hard to classify due to class mixing (g) and groups with identical dimension values (h).}
  \label{fig:use_case1}
\end{figure*}

%% file: 5.Case_Study.tex
In this section, we demonstrate how our tool can support users to better understand the general behavior of t-SNE and to validate the quality of t-SNE results by presenting a typical usage scenario and a more detailed use case, both based on data sets from the medical domain. This section follows the methodology from Ming et al.~\cite{Ming2019RuleMatrix} in order to showcase our tool's abilities to open the black box of an ML approach in a similar way.
However, the usage tasks discussed in the following are very different due to our use of the unsupervised t-SNE algorithm, in contrast to their investigations of supervised ML techniques.

\subsection{Usage Scenario: Understanding a Cancer Classifier}

Anna is a medical student who is enthusiastic about becoming a specialist in identifying and treating breast cancer. She heard about a DR algorithm called t-SNE, and she is eager to know if it can help her to identify cancer cells accurately. Personally, Anna does not completely trust the decisions made from automatic algorithms (such as classifiers), so she would prefer to use an interactive visualization. 
She decides, then, to use t-SNE to explore the Breast Cancer Wisconsin data set which she downloaded from the UCI machine learning repository~\cite{Dua2017Machine}. The data set contains measurements for 699 breast cancer cases, labeled into \emph{benign} or \emph{malignant} cancer. The nine dimensions included in this data set are cytological characteristics rated from 1 to 10 (higher means closer to malignant) when the instances were collected. However, she read on the Internet that t-SNE is a complex algorithm, and most of its decisions are hidden from the user perspective. After finding that t-viSNE allows her to interpret and assess t-SNE's results, she decides to use it.

\mypar{Overall Accuracy}
Anna loads the data into t-viSNE and starts the hyper-parameter exploration with a grid search. After the execution, she sees several projections that accurately separate the two classes. As she does not have any special preference, she selects the top-left projection, because the projections are sorted from best to worst based on the average of all the provided quality metrics. After the resulting scatterplot is loaded in the main view, she starts to investigate the overall quality by looking at the Shepard Heatmap, see \autoref{fig:use_case1}(b). Most values are situated along the diagonal of the heatmap, which---as she learned from the documentation of the tool---suggests that it is a rather accurate projection. Also, by examining the distribution of points by color in the overview (\autoref{fig:use_case1}(a)), she gets the impression that the points are mostly correctly arranged into two classes (malignant cancer cases on the left and benign cancer cases on the right). Since labels are not used by t-SNE (it is an unsupervised technique), this further supports her initial assumption that the produced results are accurate.

When she looks at the main view again, one thing catches her eye: there is quite a difference in density between the two large clusters of points (as shown by the points' colors in~\autoref{fig:use_case1}(c)). The cluster to the left (mostly \emph{malignant} cases) has low density in general, as opposed to the cluster to the right (mostly \emph{benign} cases), which seems to be quite sparse. ``That is strange,'' she thinks. It seems to be the opposite of what the t-SNE projection is showing, since the cluster to the left looks more compact in the projection. It also indicates that \emph{benign} cases are more homogeneous in the high-dimensional space, being closer to each other than the \emph{malignant} cases, and it could also mean that \emph{malignant} cases have a less clear profile.

\mypar{Interpretation of Clusters}
Anna is satisfied with her initial look at the data set through the projection, but one question comes up in her mind: how did the algorithm manage to separate the cases between benign and malignant? If she would understand how that worked, she might not only be able to validate if the results make sense, but also use that knowledge to better understand the differences between the cases in terms of cytological characteristics.
Anna uses the \emph{Dimension Correlation} in order to determine the role of the data set's dimensions in the outcome of the projection. She interactively draws a polyline with her mouse following the pattern from the benign cases to the malignant ones, as shown in~\autoref{fig:use_case1}(c). By looking at the Dimension Correlation view (see~\autoref{fig:use_case1}(d)), she observes that \cit{mitoses} is the least important dimension due to its weak correlation (approximately 18\%). She validates her hypothesis by clicking on the \cit{mitoses} dimension and observing that the actual dimension values look almost randomly distributed throughout the projected points. 
Afterward, she resets the current selection and draws two new polylines, which are perpendicular to the previous one, through the points of (1) the malignant class (see~\autoref{fig:use_case1}(e)) and (2) the benign class (not shown due to space limits). For this new investigation, she is only interested in the highest correlations, so she sets a threshold for a minimum of 20\% for a correlation to be visible. For the first case (1), it appears that t-SNE separates the malignant class according to \cit{normal nucleoli,} \cit{size uniformity,} and \cit{shape uniformity} in one area---as explained in~\autoref{fig:dim_corr}---and the other area due to \cit{bare nuclei} (\autoref{fig:use_case1}(f)). The order and direction of the produced bar charts (in accordance with the orientation of the initially-drawn shape) allowed her to reach this conclusion.
%
%
In the second case (2) (not included due to space constraints), she spotted that there is a pattern of a rapid increase in the \cit{clump thickness} (more than 80\% correlation) when going from the middle-left side to the bottom side of the cluster with the benign classified points.
``This is new,'' she thinks. These connections between the dimensions and the formation of the clusters are something she was previously not aware of. 

\mypar{Investigation of Outliers}
Next by looking back at the t-SNE overview, she identifies a red-colored instance positioned far away from the rest of the \emph{malignant} points, which grabs her attention (\autoref{fig:use_case1}(a), bottom). 
She thinks it might be an error in the projection, and decides to examine it closer by selecting a few points around the potential outlier with the lasso selection (only one point in the selection is malignant, while all others are benign). The PCP view adapts to the selection (\autoref{fig:use_case1}(g)), and she is able to acknowledge that, indeed, these points have very similar values for most dimensions, so the seemingly erroneous positioning of the point was not t-SNE's fault. ``These points are very similar, which means it must be hard to decide exactly where they belong,'' Anna thinks. She is glad she could investigate them further and check their dimensions with interactive visualization; an automatic procedure might have simply misclassified that instance, with no clear explanation of why that happened.
Finally, when zooming into the main scatterplot view, she discovers a larger number of mini-clusters, such as the one shown in~\autoref{fig:use_case1}(c), where compact points form a tight subcluster at lower zoom levels. By looking at the PCP again (\autoref{fig:use_case1}(h)), she realizes that these points are all exactly the same (i.e., they have the same dimension values). After investigating similar subclusters with a large density, she learns that t-SNE formed this and more mini-clusters in different areas of the projection as a result of their high (usually identical) similarity.


\begin{figure*}[tb]
  \centering
  \includegraphics[width=\linewidth]{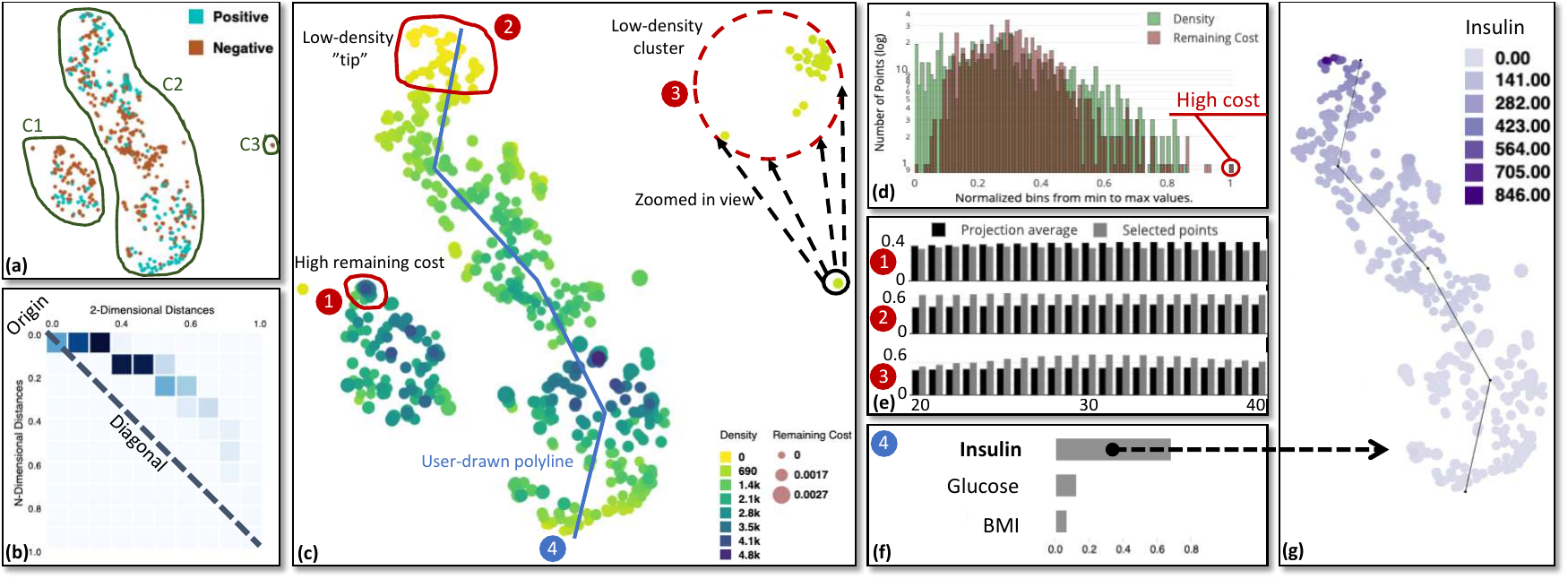}
  \caption{Use case based on the Pima Indian Diabetes data set. Although there are three separate clusters C1--C3, the class labels are mostly mixed (a), and the Shepard Heatmap (b) indicates that smaller N-D distances are spread out in 2-D. Some insights about the clusters (c): C1 has a small area with high remaining cost (d); C2 has a clearly-distorted shape that is highly correlated with the \emph{Insulin} dimension (f, g); and C3 is tight in the projection, but sparse (low density) in N-D. All (red-colored) selected areas show, in general, good Neighborhood Preservation (e) starting from $k=20$, except for the \circled{1} subcluster in C1 that decreases as $k$ increases.}
  \label{fig:use_case2}
\end{figure*}

\subsection{Use Case: Improving Diabetes Classification}

In our use case, we chose the Pima Indian Diabetes data set~\cite{Smith1988Using} to illustrate how t-viSNE can lead to a better overview, quality of the results, dimension understanding, and even performance improvements. The data set includes 768 female patients of Pima Indian heritage, aged between 21 to 81. The main task in this example is to classify the patients into positive (which have diabetes; 268 data points) or negative to diabetes (i.e., healthy; 500 data points). Every data instance contains eight dimensions: the number of times each patient/person was pregnant and their age, plasma glucose concentration level, diastolic blood pressure, skin thickness, insulin level, body mass index (BMI), and diabetes pedigree function (DPF), which is a function measuring the hereditary or genetic risk of having diabetes.

\mypar{Overall Accuracy}
We start by executing a grid search and, after a few seconds, we are presented with 25 representative projections. As we notice that the projections lack high values in \emph{continuity}, we choose to sort the projections based on this quality metric for further investigation. Next, as the projections are quite different and none of them appears to have a clear advantage over the others, we pick one with good values for all the rest of the quality metrics (i.e., greater than 40\%). The overview in~\autoref{fig:use_case2}(a) shows the selected projection with three clear clusters of varying sizes (marked with C1, C2, and C3). However, the labels seem to be mixed in all of them. That means either the projections are not very good, or the labels are simply very hard to separate. By analyzing the Shepard Heatmap (\autoref{fig:use_case2}(b)), it seems that there is a distortion in how the projection represents the original N-D distances: the darker cells of the heatmap are above the diagonal and concentrated near the origin, which means that the lowest N-D distances (up to 30\% of the maximum) have been represented in the projection with a wide range of 2-D distances (up to 60\% of the maximum). While it may be argued that the data is too \emph{spread} in the projection, we must always consider that t-SNE's goal is not to preserve all pairwise distances, but only close neighborhoods. The projection has used most of its available 2-D space to represent (as best as possible) the smallest N-D distances, which can be considered a good trade-off for this specific objective. In the following paragraphs, we concentrate on some of the goals described in Subsection~\ref{sec:neigh_pres} and Subsection~\ref{sec:dimensions} for each of the three clusters.


\mypar{C1: Remaining Cost}
Looking at the main view (\autoref{fig:use_case2}(c), \circled{1}), we detect an area on the top of cluster C1 with slightly increased size for a few points (in comparison to the other points in the same cluster), which means there are high values of remaining cost in this small area. 
%
%
This is usually a sign of a badly-optimized area that should not be trusted. To confirm that, we look at the KLD distribution (\autoref{fig:use_case2}(d)): the vast majority of points are located between $0.1$ to $0.6$ on the $x$-axis. This means that those were very well optimized (notice that the $y$-axis is in log scale). Only a handful of points show higher costs, and those few larger points in C1 belong to this group. 
%
Additionally, when we inspect the \emph{Neighborhood Preservation} plot (\autoref{fig:use_case2}(e)), we see that the badly-optimized area has lower values compared to the projection's average, but the values decrease even more after $k>26$ in contrast to those around $k=20$. That means these points are not well-positioned compared to both very close neighbors and the entire projection. These two aspects of our investigation confirm our reservations against this area.


\noindent \textbf{C2: Interpretation of Patterns} \quad
One salient pattern that stands out in the projection (\autoref{fig:use_case2}(c)) is the long curved shape of cluster C2. As opposed to C1 and C3, which look like ordinary (formless) clusters, the points in C2 have been laid out in the 2-D projection in an elongated shape going from top to bottom, with slight curves to the right and then to the left. It would be natural to hypothesize that there is some specific underlying factor in the data that caused this shape to happen, and to be curious as to what exactly that factor is. Our proposed \emph{Dimension Correlation} tool was designed to answer such questions.
For that, we first draw a polyline that simulates a \cit{skeleton} of C2's shape (\emph{user-drawn polyline} in~\autoref{fig:use_case2}(c)). The results show high correlation for the \cit{insulin} dimension along our drawn path, with a value of just below $70\%$ (\autoref{fig:use_case2}(f)), and low correlation with all other dimensions. Finally, we click on the bar to indicate that we want this specific dimension's values to be presented, which results in a clear color gradient from the bottom to the top of C2 (\autoref{fig:use_case2}(g)). This color gradient corresponds to increasing levels of insulin, as can be seen in the color legend. We can then interpret that the \emph{insulin} dimension has a high correlation with the formation of this specific shape. 

\mypar{C3: Densities}
The next step in our analysis is to confirm if the layout of the points accurately represents the original N-D densities of the clusters. By inspecting the distribution of colors over the points in the main view (\autoref{fig:use_case2}(c)), we can see that each cluster has a different density profile: C1 presents the most dense neighborhoods, C2 has average-to-high density throughout most of its points (with a small tip with very low density), and C3 has low density overall. This quick look is enough to catch two interesting insights: we confirm that the neighborhoods with highest densities (i.e., containing the smallest pairwise distances) are indeed spread out by the projection, as we had initially hypothesized from the Shepard Heatmap; and we detect a quite counter-intuitive phenomenon where the areas with the lowest density in N-D (or the most \emph{sparse} areas) are represented in 2-D in the most compact neighborhoods (marked in~\autoref{fig:use_case2}(c) as ``low-density `tip'\thinspace'' \circled{2} and \cit{low-density cluster} \circled{3}). By inspecting the \emph{Neighborhood Preservation} of the latter low-density area in C3 (\autoref{fig:use_case2}(e)), we have more evidence that this insight is indeed correct, since the small \cit{low-density cluster} \circled{3} starts with relatively high preservation from $k=20$ and becomes even larger with a peak around $k=30$. 
We can conclude that, even though this area is sparse in N-D, it presents high cohesiveness in its neighborhood, which causes t-SNE to embed the corresponding points as a compact group.\\ 


\mypar{Closing the Visual Analysis Loop}
A more detailed investigation of C3 (\autoref{fig:use_case2}(c), \circled{3}) shows that some of the internal variation of this cluster has not been well represented by t-SNE, since the points are mostly overlapping. It appears that the variation of the other clusters was prioritized by the algorithm, leaving C3 with the appearance of almost a single point. Such insights, found only through the visual analysis, also contribute to the investigation of the quality of the projection, and in t-viSNE they can be used to trigger a search for an improved projection before the visual analysis proceeds.
%
Thus, we use a lasso selection to choose C3, then use the ``optimize selection'' button (see \autoref{fig:teaser}(e), top right) to identify the best projections for the selection. After sorting the six results based on QMA, the chosen one can be seen in \autoref{fig:teaser}. The main difference between this new projection and the previously-analyzed example is that perplexity is set to 10 instead of 50, making the clusters much sparser. 
The values of all the quality metrics are still high for this new projection, 
and cluster C3 can now be entirely explored without the necessity to zoom in (cf. \autoref{fig:teaser}(f), highlighted). In \autoref{fig:teaser}(g), values of $k=1$ to $k=13$ are high, demonstrating the good Neighborhood Preservation in C3. Also, \emph{Dimension Correlation} investigation indicates that \cit{BMI} and \cit{glucose} are highly-correlated to C3 (see \autoref{fig:teaser}(j)), and \autoref{fig:teaser}(k) highlights the differences in the dimensions and the instances of C3 in connection to the better separated (compared to before) true labels, cf. \autoref{fig:teaser}(b).

%% file: 6.Evaluation.tex
In addition to the described use cases, we performed a comparative user experiment in order to gather evidence on the effectiveness of our visualization tool against another state-of-the-art alternative, Google's Embedding Projector (GEP)~\cite{Smilkov16}, as described next.
The results of a pre-study, with a single group that tested only t-viSNE, can be found in the supplemental material of this paper.
%

\subsection{Comparative User Experiment} \label{sec:exp}

The main goal of the study was to test if t-viSNE improved the usability and effectiveness of the exploration of high-dimensional data with t-SNE when compared to another state-of-the-art tool. 
Table~\ref{tab:featurescomp} in Section~\ref{sec:relwork} was used as the basis for an Analysis of Competing Hypotheses (ACH)~\cite{Heuer99}, a methodology for the fair comparison of a collection of opposing hypotheses; in our case, the multiple different views by our tool in terms of the capabilities and various possibilities they convey to the user. 
%
After the analysis, we decided on GEP mainly because it has a good overlap of functionalities with t-viSNE, is well-known, available online, and works correctly with user-provided data. \hl{VisCoDeR~\cite{Cutura18}, for example, also provides an overlap of features, but the focus of the tool and the tasks it supports---the comparison of DR methods---is very different from the focus of our experiment.} Clustervision~\cite{kwon2018}, on the other hand, did not work when we tried to load our own data sets).

\mypar{Research Questions}
The goals of the experiment are defined by two research questions, RQ1: ``Will the users spend the same time performing the tasks in both tools?'', and RQ2: ``Will both tools provide, from the users' perspective, the same level of support for the given tasks?'' Thus, we were interested in checking the \emph{completion time} for the tasks in each tool (related to RQ1). For answering RQ2, we studied the users' feedback both for specific tasks (i.e., the tool supportiveness) and in general (with the help of the ICE-T methodology, cf.~\autoref{sec:icet}).

\begin{figure}[htb]
	\includegraphics[width=\columnwidth]{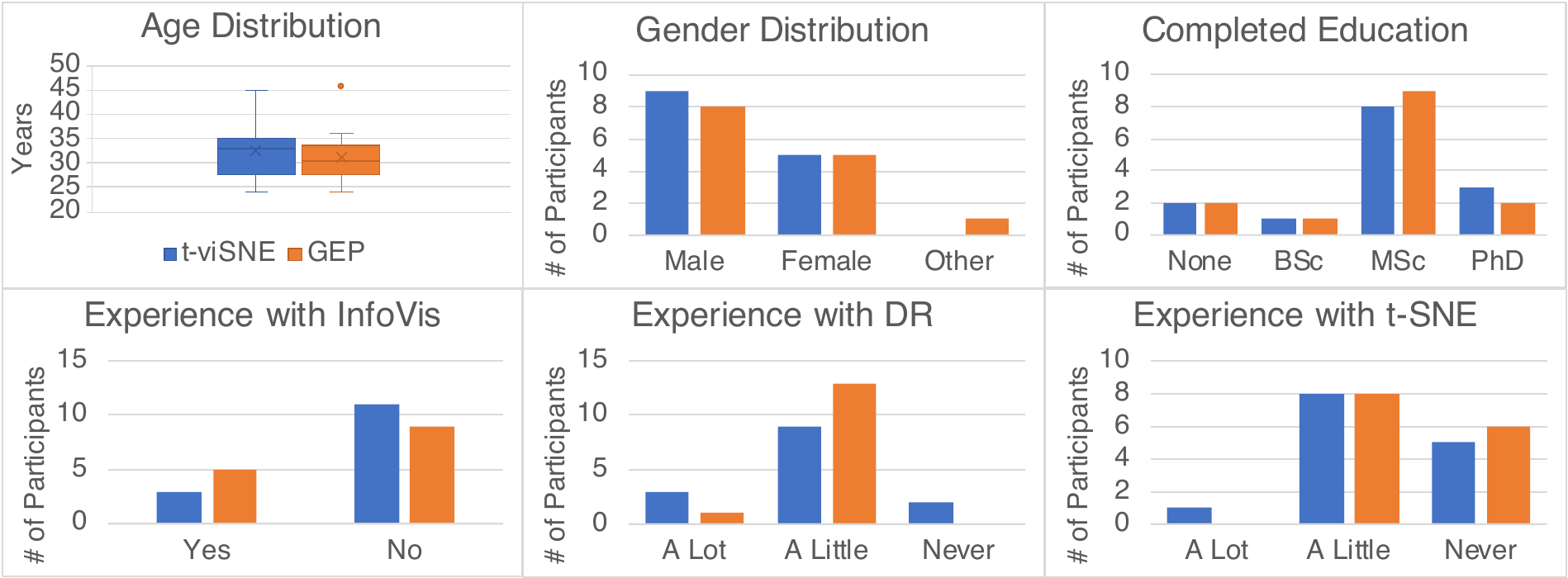}   
	\caption{Statistics on the participants of our comparative user study, split into the two groups using t-viSNE and Google's Embedding Projector (GEP).}
	\label{fig:stats}
\end{figure}

\mypar{Participants}
Our target group was data analysts who were interested in analyzing high-dimensional data, felt they needed better tools for the job, and preferably were familiar with either t-SNE or DR in general. We reached for volunteers through relevant mailing lists and contacting visualization research groups of three universities from Sweden, and the 28 respondents (19 researchers, 6 students, and 3 practitioners) were
assigned to two groups of 14 individuals: GEP and t-viSNE. The assignment was performed by preserving---as much as possible---the balance between \emph{completed education}, \emph{previous experiences}, and other characteristics, see~\autoref{fig:stats}. All of the participants except one had no color perception issues. The one who reported a minor distinction problem between almost identical shades of red and green confirmed having no problem to correctly perceive the specific color gradients when using the tool (t-viSNE). Therefore, we decided to keep those results in the study. For more details of our participants, we refer to~\autoref{fig:stats}.



\begin{figure*}[ht]
	\includegraphics[width=\textwidth]{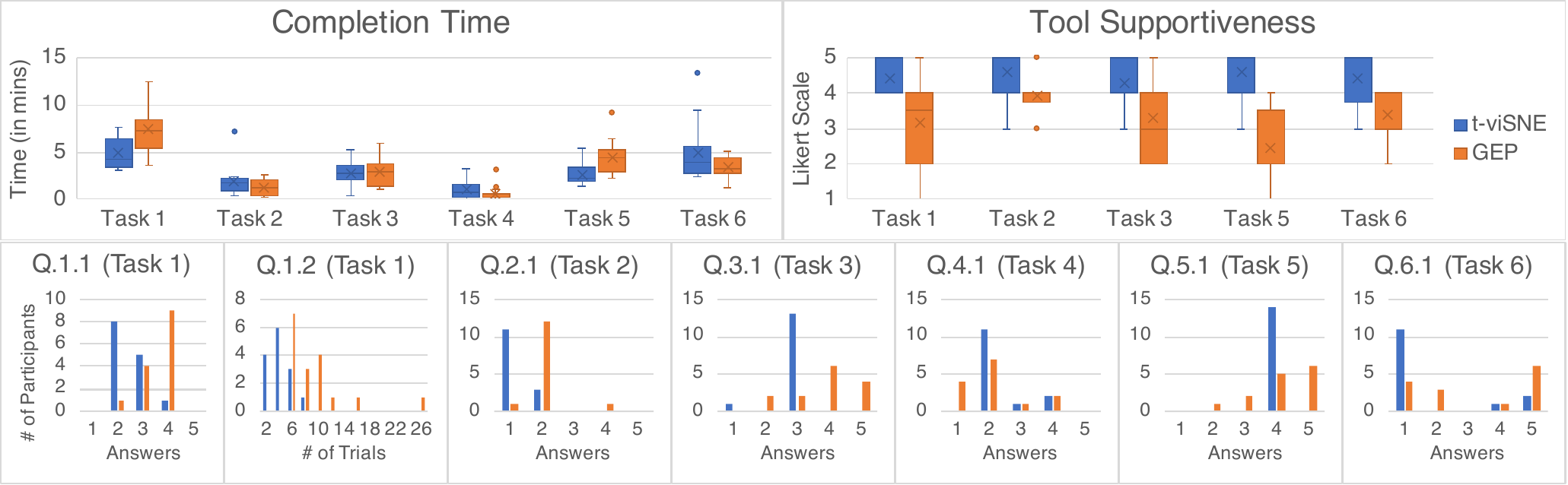} 
	\caption{Results of the comparative study: the top charts show \emph{completion time} and \emph{tool supportiveness} (as judged by participants) for all the tasks of the study, and the bottom row includes the histograms of the participants' responses in all questions/tasks. The completion times between the two groups were very similar, but t-viSNE got consistently higher scores for \emph{tool supportiveness} in all tasks. For a detailed analysis of each of the individual tasks' results, please refer to Subsection~\ref{sec:results}.}
	\label{fig:results1}
\end{figure*} 

\mypar{Study Design}
Each participant took part individually (i.e., the study was performed asynchronously for each subject, in a silent room), using the same hardware, and the study was organized into four main steps, which were identical for both groups except that each interacted with the corresponding group's \textbf{tool} (GEP or t-viSNE). First, they were shown a video tutorial which discussed t-SNE itself and the main features of the \textbf{tool} (cf. supplemental material of this work). An illustrated transcription of this tutorial was available at all times in the form of a printout. 
In the second step, after watching the video, the participants had a fixed time slot to \emph{play} with the \textbf{tool} without any specific goal, and to ask questions. After this time slot ended, no more questions were answered. 
The third step was to perform a set of specific tasks described in a handout, using a t-SNE projection of the Breast Cancer Wisconsin data set provided by the \textbf{tool}, and to answer the questions related to these tasks (see \emph{Tasks} below for details). Participants were also asked to notify when each task was completed, so we could track the task-specific completion times. 


Finally, in the fourth step, they filled out a feedback form based on the ICE-T methodology~\cite{wall2019aheuristic}.
The ICE-T evaluation form focuses on the value in and the interactivity of visualizations. It has four main high-level components, i.e., the pillars of the approach proposed by Wall et al.~\cite{wall2019aheuristic}: Insight, Confidence, Essence, and Time (ICE-T). Each of these pillars consists of two to three sub-questions representing the mid-level guidelines. Subsequently, each of these mid-level guidelines has one to three low-level heuristics, adding up to a total number of 21 heuristics in the end.  In more detail: \emph{Insight} is the ability to impel and identify insights or insightful questions about the data. \emph{Confidence} is the ability to produce confidence and trustworthiness in the data. \emph{Essence} is the capability to communicate concisely an overall essence of the data. Lastly, \emph{Time} is the capability to reduce the necessary total time to respond to a large variety of queries about data. The operationalization of this conceptual approach led to the ICE-T evaluation form, where raters can give an answer for each heuristic in a 7-point Likert rating scale, or the ``Not Applicable'' answer. 

\mypar{Tasks}
\hl{Six tasks were provided to the participants, without any specific mentions to the \textbf{tool}'s features. In consequence, the participants themselves were responsible for performing them to the best of their abilities. The six tasks were designed to match the six main pitfalls of the exploration of high-dimensional data with t-SNE, as defined by Wattenberg et al.~\cite{wattenberg2016}. Their numbering follows the same order as described in Section~\ref{intro} (so Task 1 is related to pitfall (i), Task 2 is related to pitfall (ii), and so on).} Each task consisted of one, two, or three questions that the participants were asked to answer, all with multiple choices (except for Q.1.2) including ``I do not know''. Before moving on to the next task, they were also asked to rate how supportive the \textbf{tool} was for that task.
A quick summary of the tasks is presented together with the results in the next section. Please refer to the handout provided in the supplemental materials for the complete description, as seen by the participants.


\subsection{Results} \label{sec:results}


Figure~\ref{fig:results1} provides a summary of the data gathered during the experiment, more specifically: the task completion times, the reported supportiveness of the tools on each task, and the distributions of answers to each task. The analysis of the ICE-T results can be found further below in Subsection~\ref{sec:icet}.

One initial observation is that the overall \emph{Completion Time} for both groups was remarkably similar. With the exception of Tasks 1 and 5, where t-viSNE users performed faster than GEP, in general the results have not shown any statistically significant difference. To answer RQ1, \textbf{we detected no statistically significant difference in the time the users needed to perform the given tasks for both tools, in general}. 

On the other hand, t-viSNE obtained consistently higher scores for \emph{Tool Supportiveness}, with a higher average in all the proposed tasks. The bulk of the distributions of the supportiveness scores from the two groups overlap little, mostly near outliers (the ``N/A'' option was chosen three times, all in the GEP group).
While this is of course based on subjective user feedback, we consider that it is nonetheless an important aspect of the results; since both t-viSNE and GEP mainly aim to support the \emph{exploratory} visual analysis of high-dimensional data---through many different coordinated views and interactive tools---it may become hard to set a single, concrete ground-truth for evaluating their perfomance as a whole. Thus, the users' perception of how much the tool supported their intended goals can be one (but not the only) good indication of how useful the tool actually is.

\mypar{Task-Specific Qualitative Analysis}
We proceed by comparing the results of the two groups in each task individually, using the task-specific histograms from the bottom row of Figure~\ref{fig:results1}. Our goal here is to perform an informal and qualitative analysis of the results, using the data from the experiment as input, to obtain more insights on the differences of the user experiences with the two different tools.

In Task 1, \emph{Choosing More Effective Parameters}, participants were asked to find problems in their chosen t-SNE layout (Q.1.1) and to note how many times they tuned the t-SNE parameters before starting the experiment (Q.1.2). In both cases, smaller is better (i.e., fewer problems and easier to tune the parameters).
For Q.1.1, the distributions of the answers of the two groups are symmetrically opposite: most t-viSNE users found fewer problems with the initial layout (answers 2 and 3), while most GEP users found many problems, to the point of considering them too hard to count (i.e., answers 3 and 4). This indicates that t-viSNE users were, in general, more satisfied with the t-SNE layout after setting the parameters.
The answers to Q.1.2 also show that t-viSNE users needed fewer iterations to find a good parameter setting.

In Task 2, \emph{Deciding About (Ir-)Relevant Sizes of Clusters}, the goal was to determine the relative density (or, conversely, the sparsity) of the clusters. The expected answer---see the visualization in Figure~\ref{fig:use_case1}(c), for example---is that the benign cluster is denser (even though it may appear less dense, when no extra information is provided in the projection), which corresponds to answer 1. We can see from the histogram of Q.2.1 that most of the t-viSNE group agreed with this result, while the GEP group mostly chose answer 2: ``The benign cluster is sparser than the malignant''. \pagebreak

For Task 3, \emph{Evaluating Original Space Distances}, participants had to judge the quality of the distance preservation in the projection. Most participants from the t-viSNE group chose answer 3---good (but not perfect) distance preservation---which seems to align well with the Shepard Heatmap from Figure~\ref{fig:use_case1}(b), for example. The answers from the GEP group were mostly scattered, with a tendency towards answers 4 (distances are only slightly preserved) or 5 (``I do not know'').

Task 4, \emph{Extracting Patterns from the Projection}, consisted simply of determining the number of clusters in the projection. The results from both groups were quite similarly distributed, with most participants choosing 2 clusters (as expected, see e.g. Figure~\ref{fig:use_case1}(a)). One difference is that 4 participants from the GEP group chose 1 cluster, which could indicate that GEP failed to clearly separate the two clusters in some cases.

For Task 5, \emph{Observing and Exploring Shapes}, participants were asked to determine the least important dimension that affected the shape of the clusters. All participants from the t-viSNE group chose answer 4, \emph{mitoses}, in agreement with our own observations for this data set (e.g., Figure~\ref{fig:use_case1}(d)) and previous work (e.g.,~\cite{borges2015}). While we cannot claim that this is the \emph{correct} answer, the results are encouraging from the perspective of the consistency of the participants' experience with the tool. The GEP answers, on the other hand, were mostly scattered, but with a tendency towards answer 5 (``I do not know'').


Finally, the goal of Task 6, \emph{Interpreting and Assessing Local Topology}, was to find and interpret ``unusual'' patterns in the projection, more specifically formations that are known to happen in this data set because of identical points, i.e., data points which have the same values for all dimensions. This corresponded to answer 1, which was correctly identified by most participants from the t-viSNE group. The answers for the GEP group were again mostly scattered, with 6 of them choosing ``I do not know'' (against 2 only from the t-viSNE group).


\subsection{ICE-T Results} \label{sec:icet}

As described in Subsection~\ref{sec:exp}, we complemented the data from the tasks themselves by using the ICE-T methodology and questionnaire to gather and compare structured user feedback from both groups. The scores obtained from all participants, for all ICE-T components, can be seen in Table~\ref{tab:icet}. Larger is better, with green indicating good results (as opposed to red). The raw data is accompanied by two statistical analyses: the two-tailed 95\% confidence intervals (CIs) per component ($t^* = 2.16, N=14$); and the results of one-tailed Mann-Whitney U tests~\cite{corder14}, also one per component, with a significance level of 0.01 ($U^* = 47$). We chose a non-parametric test due to the small sample size and its robustness to non-normality in the data distribution.

A quick visual inspection of the two tables already hints at t-viSNE having superior scores than GEP in all components, with all cells being green-colored (as opposed to GEP's table, which contains many red-colored cells). Indeed, the smallest score for t-viSNE was 4.75, while GEP got many scores under 4 (or even under 3). Following the trend of the previously-presented results, the \emph{Time} component is the one with the most similar scores between the two tools. On the other hand, the \emph{Confidence} component had the largest difference, which suggests that participants were significantly more confident in their results when using t-viSNE than with GEP.
The observed conclusions are confirmed when we compare the component-wise CIs for both groups---since none of them overlap---and the results of all component-wise Mann-Whitney U tests, with all U's well below the critical value of 47, showing that t-viSNE had significantly larger scores in all four ICE-T components. These results, together with the tools' supportiveness outcomes (discussed above in Subsection~\ref{sec:results}), suggest that \textbf{our tool provides better level of support for the given tasks than GEP}, which answers RQ2.

\begin{table}[t]
	\caption{Results from the ICE-T feedback. t-viSNE obtained significantly larger scores than Google's Embedding Projector (GEP) in all components.}\vspace{-2mm}
	\includegraphics[width=\columnwidth]{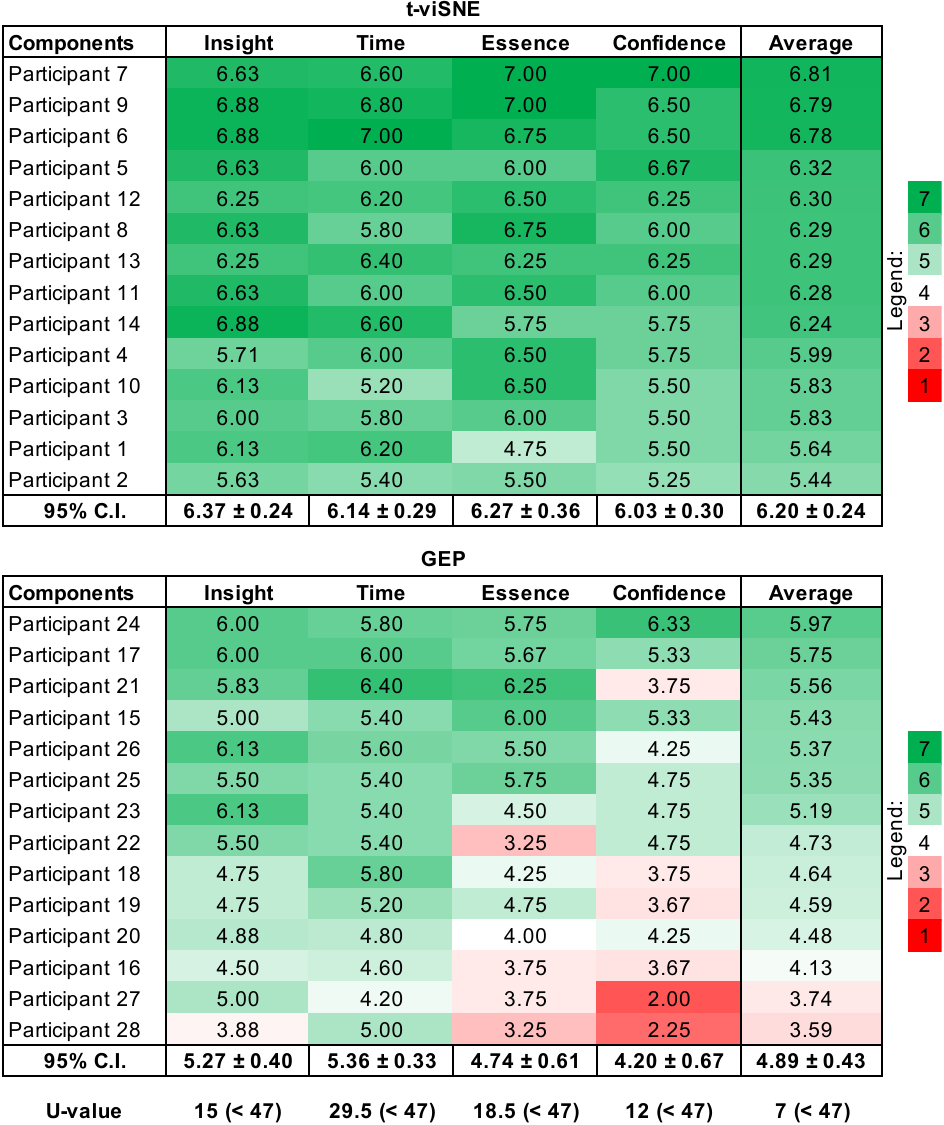}   
	\label{tab:icet}
\end{table}

%% file: 7.Discussion.tex
In this section, we discuss different aspects of the design choices of our implementation, elaborate on our experiences with developing t-viSNE, and lastly, we present limitations and future work.


\subsection{Design Choices} \label{sec:design}

\begin{figure}[htb]
  \centering
  \includegraphics[width=\linewidth]{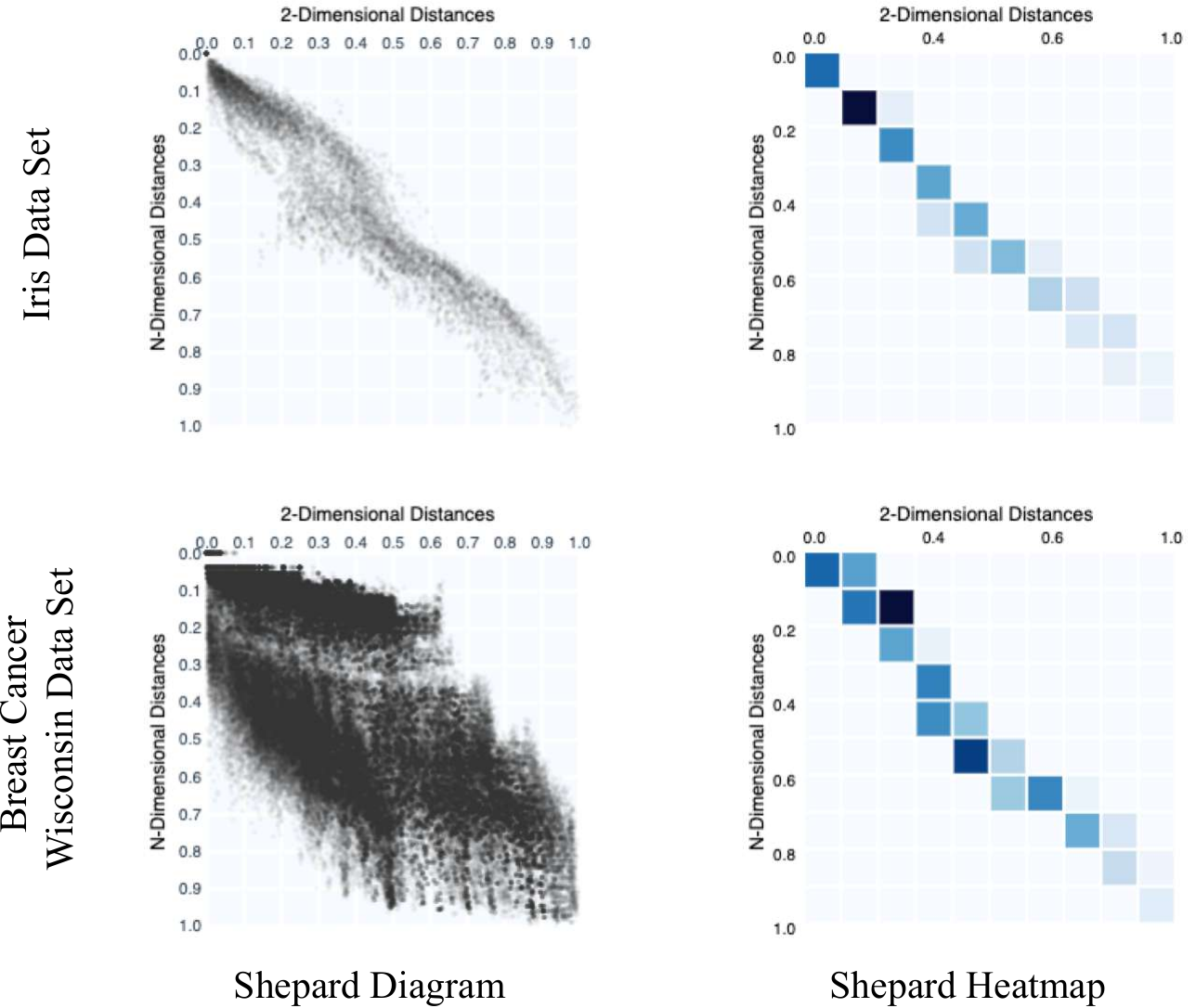}
  \caption{Comparison: Shepard Heatmap vs. Shepard Diagram.}
  \label{fig:shepardcomp}
\end{figure}

\mypar{Shepard Heatmap vs. Shepard Diagram}
We propose the Shepard Heatmap, instead of simply adopting a Shepard Diagram as usual in previous work, in order to make sure this view reaches its intended goal: to be a quick and simple overview of the quality. A full scatterplot with ${(n^2-n)}/{2}$ points and variable transparency would have done a similar job when it comes to avoiding clutter, but that would mean (a) a lot of unnecessary details, such as outliers, would be visible and might attract the user's attention, and (b) t-viSNE would show several scatterplots at the same time, which could be confusing for the user. During our design process, we realized that a different abstraction, with less detail, was the superior choice based on the hypothesis that grid-based binning can reduce cluttering and overlapping~\cite{Longshaw08}, while hiding some of the less-prominent details. In~\autoref{fig:shepardcomp}, we show two examples of the results of this trade-off: for the smaller Iris data set, both diagrams seem to convey the same patterns, but for the somewhat larger Breast Cancer Wisconsin data set (described in~\autoref{case}), the patterns are more confusing while using a Shepard Diagram.
\hl{We decided to implement both approaches in our tool, as shown in~\autoref{fig:teaser}(c), so that the user may choose to fall back to the more common scatterplot-based view if desired.}
%
Additionally, as for the bin sizes of the heatmap, we decided to keep them constant (with 10 bins by default) in order to make sure that every projection can be interpreted in a predictable way, without extra training or parameter settings required from the users. The color scale of the heatmap adapts automatically to the range of distances of the loaded data set, divided into 10 discrete sub-ranges. Implementing both bin sizes (grid and color) as user-defined parameters would be a trivial addition to the tool.

\mypar{Different Colormaps}
There are quite a few different colormaps being used simultaneously in t-viSNE: as a bare minimum, there is a categorical one for the labels in the overview (and the PCP), a single-hue sequential one for the Shepard Heatmap, and a multi-hue sequential one for the main view. We carefully chose these colormaps, considering Gestalt laws and recent research results~\cite{Liu2018}, in order to make sure they are efficient, do not interfere with each other, and that it is as clear as possible that they represent different things. 

\begin{figure}[htb]
  \centering
  \includegraphics[width=\linewidth]{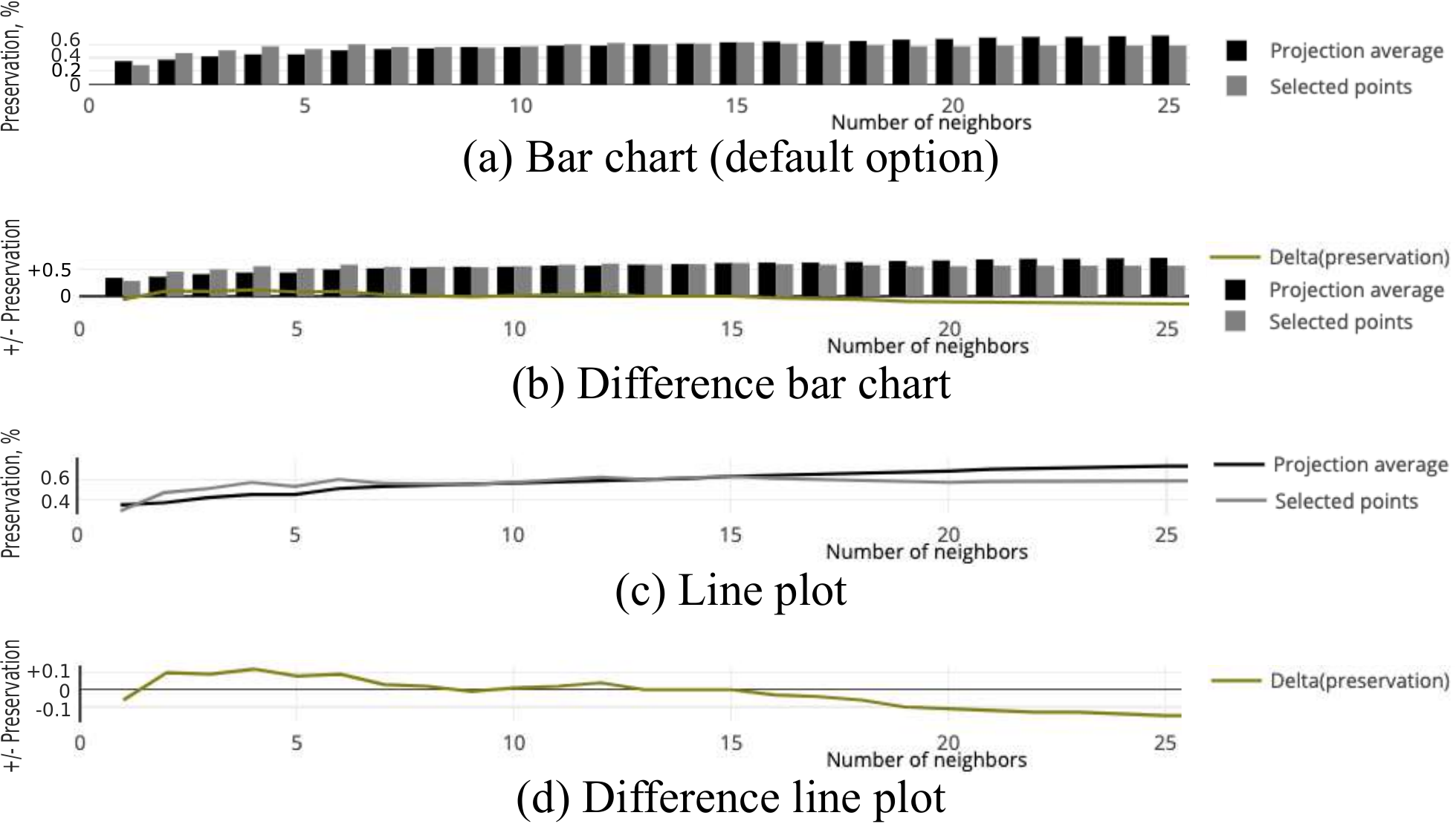}
  \caption{Four options for the visualization of Neighborhood Preservation (using the Iris data set).}
  \label{fig:neighDesign}
\end{figure}

\mypar{Visual Abstraction for Neighborhood Preservation}
The Neighborhood Preservation plot (\autoref{fig:teaser}(g)) can be visualized as a bar chart (by default), a difference bar chart, a standard line plot, or a difference line plot, as shown in~\autoref{fig:neighDesign}. Although they show basically the same information, each one has advantages and disadvantages. On the one hand, we found the bar chart (a) to be better when comparing the projection's average with the selection's average when we search for discrete k-values, and during the initial state (no selection of points), where the user can easily distinguish the bars having the same size.
It can optionally be replaced by the line plot (c), with similar effects; however, it can become confusing when there is very little difference between the selection and the projection average, due to the overlap of the two lines.
%
The difference line plot (d), on the other hand, builds on the standard plot by highlighting the differences between the selection and the global average, shown as positive and negative values around the 0 value of the y-axis. 
It provides a clearer overall picture of the difference in preservation among all the shown scales, but compromises the precision and simplicity of interpretation of the y-axis (where the exact percentage of Neighborhood Preservation was previously shown). The difference bar chart (b) is a combination of the designs (a) and (d). Similar to (d), the interpretation of the y-values might be misleading.
Lacking a clear winner in this case, we opted to let the users decide.


\mypar{Adaptive PCP vs. PCP}
Although it is not uncommon to find tools that use PCP views together with DR-based scatterplots (e.g., iPCA~\cite{jeong09}) with various schemes for re-ordering and prioritizing the axes (e.g.,~\cite{Ankerst98,Lu16}), the arrangement and presentation of these PCP's are usually static in order to reflect attributes of the data (or the projection) as a whole. In our proposed \emph{Adaptive PCP}, the arrangement of the axes is dynamically updated every time the user makes a new selection (using a local PCA); this way, the PCP only shows, at any given time, the most relevant dimensions for the user's current focus, which may differ significantly from the global aspects of the projection as a whole. Coupled with the Dimension Correlation view, this provides a highly-customized toolset for inspecting and interpreting the meanings of specific neighborhoods of data points.

\begin{figure}[htb]
  \centering
  \includegraphics[width=\linewidth]{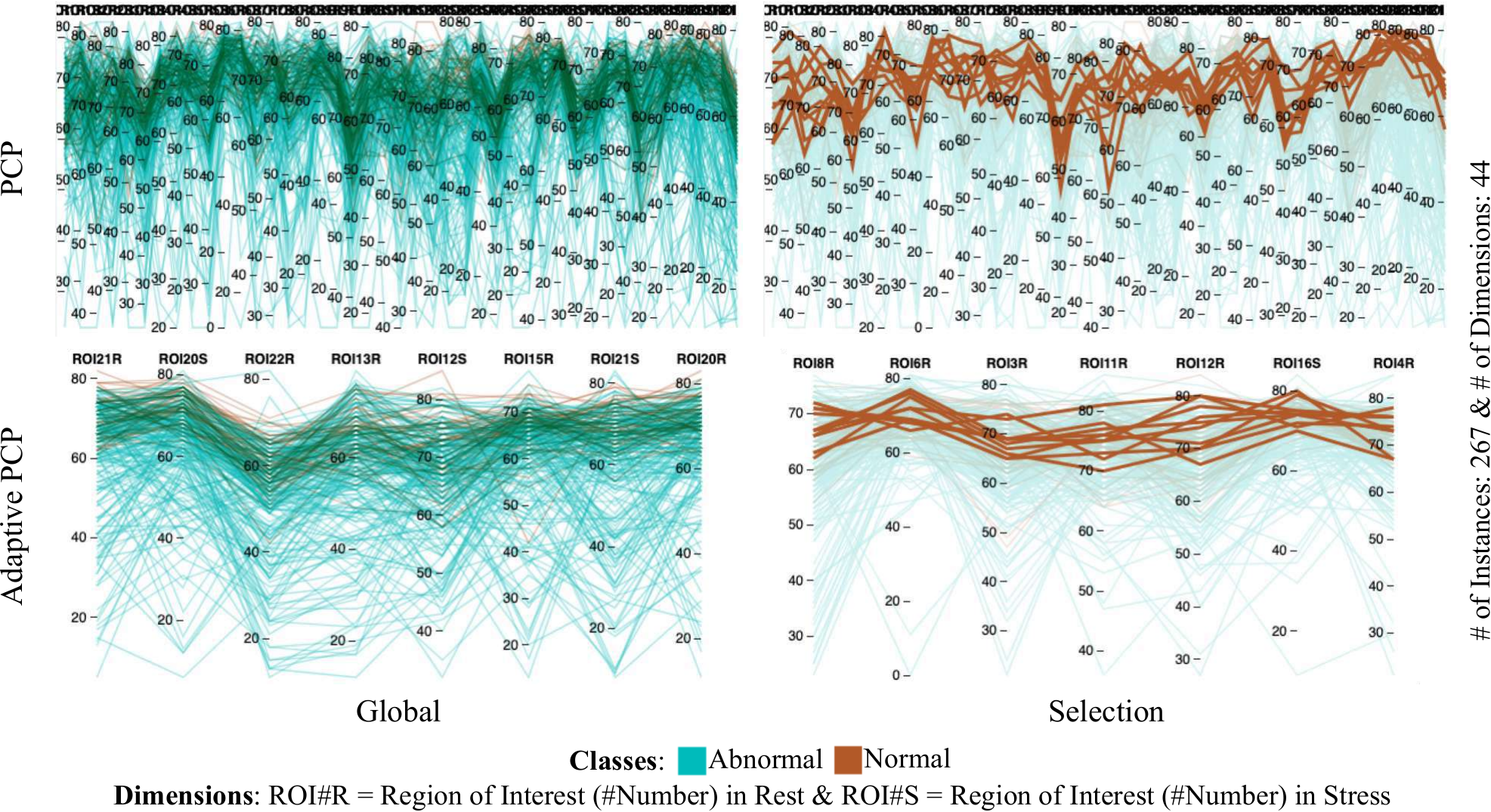}
  \caption{Adaptive PCP vs. PCP on the SPECTF data set. We demonstrate two cases: without selection of points (left) and with selection of ten points all belonging to the \emph{normal} class (right).}
  \label{fig:apcp}
\end{figure}

To briefly present the benefits of using our technique, we employ the Single Proton Emission Computed Tomography (SPECTF) data set~\cite{Dua2017Machine} with 44 dimensions. In~\autoref{fig:apcp}, we can observe that the standard PCP is cluttered, especially for the case without any selection. Thus, it is hard to see why the \emph{normal} class is actually separated from the \emph{abnormal} one. 
Furthermore, the numerous axis labels introduce even further cluttering and confusion for the users of the standard PCP. Instead, our Adaptive PCP utilizes PCA as a \emph{degree-of-interest} function and only displays the 8 most informative dimensions. It enables the analyst to discover that \emph{abnormal} classified patients have less fluctuating measurements than the others, which becomes even more salient in the selection case where the measurements for the \emph{normal} class (in brown color) are rather stable when patients are in both rest and stress conditions.

\mypar{Labels}
In order to better explain the contribution of t-viSNE, the data sets used in our use cases contain predefined labels, which is not the case in general when using unsupervised learning techniques, such as t-SNE. There is no restriction, however, to having labels when using t-viSNE; one might use the results of a clustering algorithm, for example, as a replacement for pre-defined labels, or simply no labels at all. Apart from not having any specific color mapping in the overview and the PCP, none of the other techniques are affected by it.

\subsection{Limitations and Future Work} \label{sec:limit}

We implemented t-viSNE in JavaScript and WebGL, using a combination of D3.js~\cite{D3}, Three.js~\cite{threejs}, and Plotly.js~\cite{plotly} for the frontend. In the backend, it uses Laurens van der Maaten's Barnes-Hut t-SNE implementation written in Python and C++~\cite{VanDerMaaten14}, and Projlib~\cite{projlib} for the quality measures. The use cases and experiments were performed on a MacBook Pro 2018 with a 2.8 GHz Intel Core i7 CPU, a Radeon Pro 555 2048 MB GPU, 16 GB of RAM, and running macOS Mojave.

\mypar{Performance}
There are two reasons why we decided to use the Barnes-Hut implementation of the original t-SNE algorithm~\cite{VanDerMaaten14}, instead of a newer and faster implementation~\cite{pezzotti17,chan18}. First, each fast and approximated implementation of t-SNE introduces its own variations to the algorithm, and we did not want these variations to influence the design of our tool or introduce unnecessary bias in the results of our study. Second, in this phase of the research, we were mainly concerned with designing and validating the system with the right set of views and the right analysis workflow, so we decided to prioritize the ease of implementation over the raw performance.
Replacing the actual implementation of t-SNE should be straightforward, if deemed necessary.

\mypar{Other DR Methods}
Although our main design goal was to support the investigation of t-SNE projections, most of our views and interaction techniques are not strictly confined to the t-SNE algorithm. For example, the Dimension Correlation view could, in theory, be applied to any projection generated by any other algorithm. Its motivation, however, came from the fact that t-SNE is especially known to generate hard-to-interpret shapes in its output~\cite{wattenberg2016}, so the necessity of exploring and investigating such shapes became more apparent than with other DR methods. The same goes for other views, such as Neighborhood Preservation or Adaptive PCP: the inspiration and the design constraints came from known shortcomings and characteristics of t-SNE, such as its focus on optimizing neighborhoods of points in detriment of global distances, but the implementation could be re-used in different scenarios. 
The analysis of density, however, is one example of an inherent characteristic of t-SNE, since it comes directly from its algorithm. A limitation that arises from building a tool that is tuned to tackle problems concerning a particular algorithm is the possibility of the algorithm becoming obsolete or being replaced by a newer, better alternative. We argue, though, that more than a decade after its proposal, it has now become quite clear that t-SNE is not going away anytime soon. Papers are still regularly coming out proving its stability~\cite{bodt18,bodt18b,linderman19}, and high-impact applications and publications in many different domains geared towards non-visualization and non-ML experts are based on it~\cite{vanUnen17,linderman19b}.
%
Even in the improbable scenario that t-SNE becomes obsolete soon, the fact that most of our proposed views can be re-used or adapted to different DR methods means that our work is still relevant and largely future-proof.

\mypar{User Study}
The goals of the comparative study presented in this paper were to provide initial evidence of the acceptance of t-viSNE by analysts, the consistency of their results when exploring a t-SNE projection using our tool, and the improvement over another state-of-the-art tool.
The tasks of the study were designed to test how each tool helps the analyst in overcoming the six pitfalls defined by Wattenberg et al.~\cite{wattenberg2016}), which was also one of the design goals of t-viSNE itself. Since that might not have been the case for GEP, this could be seen as a bias towards t-viSNE.
%
Nevertheless, while it may not reflect reality in the same way as, e.g., a large-scale field study performed with real-world experts in their actual working environment~\cite{carpendale2008}, the positive results from the study showed that our approach is promising and deserves to be developed and tested further, which will be done in future work.

\mypar{Progressive Quality Analysis}
The remaining costs are one aspect of estimating the projection quality. This means that projected points with high remaining costs can be moved by an additional optimization step. Akin to this idea, t-viSNE might show a preview of the data points in the next optimization step. In consequence, users could determine whether the t-SNE optimization is completed or not, simply by observing the points' trajectories in low-dimensional space. This remains as possible future work.


%% file: 8.Conclusion.tex
In this paper, we introduced t-viSNE, an interactive tool for the visual investigation of t-SNE projections. By partly opening the black box of the t-SNE algorithm, we managed to give power to users allowing them to test the quality of the projections and understand the rationale behind the choices of the algorithm when forming clusters. Additionally, we brought into light the usually lost information from the inner parts of the algorithm such as densities of points and highlighted areas which are not well-optimized according to t-SNE. 
To confirm the effectiveness of t-viSNE, we presented a hypothetical usage scenario and a use case with real-world data sets. We also evaluated our approach with a user study by comparing it with Google's Embedding Projector (GEP): the results show that, in general, the participants could manage to reach the intended analysis tasks even with limited training, and their feedback indicates that t-viSNE reached a better level of support for the given tasks than GEP. However, both tools were similar with respect to completion time.